%% file: paper.tex
\documentclass[letterpaper, 10 pt, conference]{ieeeconf}

\IEEEoverridecommandlockouts %

\usepackage{amsmath,amsfonts,amssymb}
\usepackage{amsfonts} %
\usepackage{mathrsfs} %
\usepackage{bbold} %
\usepackage{pgfplots}
\usepackage[ruled, vlined, linesnumbered]{algorithm2e}
\usepackage[tight]{subfigure}
\usepackage{tabularx}
\usepackage{multirow}
\usepackage{siunitx}

\usepackage[export]{adjustbox}

\usepackage{caption}
\usepackage{xargs} %

\usepackage[style=ieee, backend=biber, maxbibnames=5, mincitenames=1, maxcitenames=2, url=false, doi=false, isbn=false, citestyle=numeric-comp, natbib=true]{biblatex}

\usepackage[colorinlistoftodos,prependcaption,textsize=small]{todonotes}

\newcommandx{\change}[2][1=]{\todo[inline,linecolor=blue,backgroundcolor=blue!25,bordercolor=blue,#1]{#2}}

\usepackage{bm} %
\usepackage{lipsum}
\newcommand{\Secref}[1]{Section~\ref{#1}}

\newcommand\copyrighttext{%
	\scriptsize \textcolor{blue}{\textcopyright 2018 IEEE. Personal use of this material is permitted.  Permission from IEEE must be obtained for all other uses, in any current or future media, including reprinting/republishing this material for advertising or promotional purposes, creating new collective works, for resale or redistribution to servers or lists, or reuse of any copyrighted component of this work in other works}}
\newcommand\copyrightnotice{%
	\begin{tikzpicture}[remember picture,overlay]
		\node[anchor=north,yshift=-7.5pt] at (current page.north) {\fbox{\parbox{\dimexpr\textwidth-\fboxsep-\fboxrule\relax}{\copyrighttext}}};
	\end{tikzpicture}%
}

\pgfplotsset{compat=newest}
\pgfplotsset{plot coordinates/math parser=false}
\usetikzlibrary{decorations.markings, positioning}

\newlength\figureheight 
\newlength\figurewidth 

\title{\LARGE \bf Spatiotemporal Motion Planning with Combinatorial Reasoning for Autonomous Driving}
\author{Klemens Esterle$^{1}$, Patrick Hart$^{1}$, Julian Bernhard$^{1}$ and Alois Knoll$^{2}$%
	\thanks{$^{1}$Klemens Esterle, Patrick Hart and Julian Bernhard are with fortiss GmbH, An-Institut Technische Universit\"{a}t M\"{u}nchen, Munich, Germany}%
	\thanks{$^{2}$Alois Knoll is with Robotics, Artificial Intelligence and Real-time Systems, Technische Universit\"{a}t M\"{u}nchen, Munich, Germany}%
}

\begin{document}

\maketitle
\copyrightnotice
\thispagestyle{empty}
\pagestyle{empty}

\global\csname @topnum\endcsname 0
\global\csname @botnum\endcsname 0

\newcommand{\figurename}{Fig. }

\input{commands}

\begin{abstract}
\input{abstract}
\end{abstract}

\IEEEpeerreviewmaketitle

\input{introduction}

\input{problem}

\input{method}

\input{evaluation}

\input{conclusion}
\renewcommand{\bibfont}{\small}
\printbibliography

\appendix
\input{appendix}

\end{document}

%% file: commands.tex
\newcommand {\vect} {\boldsymbol}
\newcommand {\matr} {\boldsymbol}

\newcommand{\state} {\vect{x}}
\newcommand{\stateSpace} {\vect{\mathcal{X}}}
\newcommand{\beliefstate} {\vect{b}}
\newcommand{\contr} {\vect{u}}
\newcommand{\contrSpace} {\vect{\mathcal{U}}}
\newcommand{\meas} {\vect{y}}
\newcommand{\procNoise}{\vect{w}}

\newcommand {\cov}  {\matr{\Sigma}}

\newcommand{\stateNoDelta}{\hat\state}
\newcommand{\contrNoDelta}{\hat\contr}
\newcommand{\procNoiseNoDelta}{\hat\procNoise}

\newcommand{\abc}[2][\empty]{%
  \ifthenelse{\equal{#1}{\empty}}
    {no opt, mand.: \textbf{#2}}
    {opt: \textbf{#1}, mand.: \textbf{#2}}
}

\newcommand {\noiseu} {\procNoise}
\newcommand {\covu} {\matr{\Sigma_{\noiseu,}}}

\newcommand {\noiseuNoDelta} {\procNoiseNoDelta}
\newcommand {\covuNoDelta} {\matr{\Sigma_{\noiseuNoDelta,}}}

\newcommand {\defnoiseu}[1][\empty]{
 \ifthenelse{\equal{#1}{\empty}}
    {\noiseu\sim N(0,\covu)}
    {\noiseu_{#1}\sim N(0,\covu_{#1})}
}

\newcommand {\defnoiseuNoDelta}[1][\empty]{
 \ifthenelse{\equal{#1}{\empty}}
    {\noiseuNoDelta\sim N(0,\covuNoDelta)}
    {\noiseuNoDelta_{#1}\sim N(0,\covuNoDelta_{#1})}
}

\newcommand {\covm} {\matr{R}}
\newcommand {\noisem} {\vect{\nu}}
\newcommand {\defnoisem}[1][\empty]{
 \ifthenelse{\equal{#1}{\empty}}
    {\noisem\sim N(0,\covm)}
    {\noisem_{#1}\sim N(0,\covm_{#1})}
}

\newcommand{\stateB}{\vect{\xi}}
\newcommand{\contrB}{\vect{\nu}}
\newcommand{\procNoiseB}{\vect{\omega}}

\newcommand{\AB}{\mathcal{A}}
\newcommand{\BB}{\mathcal{B}}
\newcommand{\WB}{\mathcal{W}}
\newcommand{\costStateB}{\mathcal{Q}}
\newcommand{\costContrB}{\mathcal{R}}
\newcommand{\covStatesB}{\mathcal{S}_{\state}}
\newcommand{\covProcNoiseB}{\mathcal{S}_{\procNoise}}

\newcommand{\FB}{\mathcal{F}}

\newcommand{\cct}{\vect{t}}
\newcommand{\ccsval}{s}
\newcommand{\ccT}{\matr{T}}
\newcommand{\ccsvec}{\vect{s}}

\newcommand{\costState}{\matr{Q}}
\newcommand{\costContr}{\matr{R}}
\newcommand{\feedbackMatrix}{\matr{K}}
\newcommand{\cost}{J}

\newcommand{\stateConstraintMatrix}{\matr{C}}
\newcommand{\stateConstraintVector}{\vect{c}}

\newcommand{\stateConstraintFunc}{c}

\newcommand{\contrConstraintMatrix}{\matr{D}}
\newcommand{\contrConstraintVector}{\vect{d}}

\newcommand{\contrConstraintFunc}{d}

\newcommand{\stateRef}{\state^{*}}
\newcommand{\contrRef}{\contr^{*}}

\newcommand{\stateDelta}{\Delta\state}
\newcommand{\contrDelta}{\Delta\contr}

\newcommand {\Comment}[1]{\textcolor{blue}{#1}}

\newcommand {\partialder}[4][\bigg]{\frac{\partial #2}{\partial #3}#1|_{#4}}
\newcommand {\partialdernoarg}[3][\bigg]{\frac{\partial #2}{\partial #3}#1}

\newcommand{\nat}{\mathbb{N}}
\newcommand{\real}{\mathbb{R}}
\newcommand{\compl}{\mathbb{C}}

\newcommand{\norm}[1]{\left\| #1 \right\|}

\newcommand{\half}{\frac{1}{2}}

\newcommand{\parenth}[1]{ \left( #1 \right) }
\newcommand{\bracket}[1]{ \left[ #1 \right] }
\newcommand{\accolade}[1]{ \left\{ #1 \right\} }
\newcommand{\pardevS}[2]{ \delta_{#1} f(#2) }
\newcommand{\pardevF}[2]{ \frac{\partial #1}{\partial #2} }

\newcommand{\vecii}[2]{\begin{pmatrix} #1 \\ #2 \end{pmatrix}}
\newcommand{\veciii}[3]{\begin{pmatrix}  #1 \\ #2 \\ #3	\end{pmatrix} }
\newcommand{\veciv}[4]{\begin{pmatrix}  #1 \\ #2 \\ #3 \\ #4	\end{pmatrix}}

\newcommand{\matii}[4]{\left[ \begin{array}[h]{cc} #1 & #2 \\ #3 & #4 \end{array} \right]}
\newcommand{\matiii}[9]{\left[ \begin{array}[h]{ccc} #1 & #2 & #3 \\ #4 & #5 & #6 \\ #7 & #8 & #9	\end{array} \right]}

\newcommand{\transp}{^{\intercal}}
\newcommand{\Reg}{$^{\textregistered}$}
\newcommand{\reg}{$^{\textregistered}$ }
\newcommand{\Tm}{\texttrademark}
\newcommand{\tm}{\texttrademark~}
\newcommand {\bsl} {$\backslash$}

\newtheorem{theorem}{Theorem}[section]
\newtheorem{lemma}[theorem]{Lemma}
\newtheorem{corollary}[theorem]{Corollary}
\newtheorem{remark}[theorem]{Remark}
\newtheorem{definition}[theorem]{Definition}
\newtheorem{equat}[theorem]{Equation}
\newtheorem{example}[theorem]{Example}
\newcommand{\insertfigure}[4]{ %
	\begin{figure}[htbp]
		\begin{center}
			\includegraphics[width=#4\textwidth]{#1}
		\end{center}
		\vspace{-0.4cm}
		\caption{#2}
		\label{#3}
	\end{figure}
}

\newcommand{\refFigure}[1]{\figurename \ref{#1}}
\newcommand{\refChapter}[1]{chapter \ref{#1}}
\newcommand{\refSection}[1]{section \ref{#1}}
\newcommand{\refParagraph}[1]{paragraph \ref{#1}}
\newcommand{\refEquation}[1]{(\ref{#1})}
\newcommand{\refTable}[1]{Table \ref{#1}}
\newcommand{\refAlgorithm}[1]{Algorithm \ref{#1}}

\newcommand{\rigidTransform}[2]
{
	${}^{#2}\!\mathbf{H}_{#1}$
}

\newcommand{\code}[1]
 {\texttt{#1}}

\newcommand{\comment}[1]{\marginpar{\raggedright \noindent \footnotesize {\sl #1} }}

\newcommand{\clearemptydoublepage}{%
  \ifthenelse{\boolean{@twoside}}{\newpage{\pagestyle{empty}\cleardoublepage}}%
  {\clearpage}}

\newcommand{\etAl}{\emph{et al.}\mbox{ }}

\newcommand{\todoi}[1]{\todo[inline]{#1}}

%% file: abstract.tex
Motion planning for urban environments with numerous moving agents can be viewed as a combinatorial problem. With passing an obstacle before, after, right or left, there are multiple options an autonomous vehicle could choose to execute. These combinatorial aspects need to be taken into account in the planning framework.
We address this problem by proposing a novel planning approach that combines trajectory planning and maneuver reasoning. We define a classification for dynamic obstacles along a reference curve that allows us to extract tactical decision sequences. We separate longitudinal and lateral movement to speed up the optimization-based trajectory planning. 
To map the set of obtained trajectories to maneuver variants, we define a semantic language to describe them.
This allows us to choose an optimal trajectory while also ensuring maneuver consistency over time. We demonstrate the capabilities of our approach for a scenario that is still widely considered to be challenging.

%% file: introduction.tex
\section{Introduction}
\label{sec:introduction}

\subsection{Motivation}
\label{subsec:motivation}
Autonomous driving intends to relieve the driver of the task of driving, thus promising great improvements in terms of safety and comfort. With encouraging solutions for the perception task enabled by deep learning, the behavior generation remains one of the biggest challenges for autonomous driving in order to achieve full autonomy. The behavior generation problem is to find an optimal motion regarding safety and comfort under the premise of obeying traffic rules, vehicle kinematics and dynamics. Satisfying real-time demands to ensure reactiveness to dynamic obstacles in critical scenarios is a key challenge for all motion planning algorithms \cite{Gonzalez2016}.

\begin{figure}[t]
	\centering
		\includegraphics[width=0.8\columnwidth]{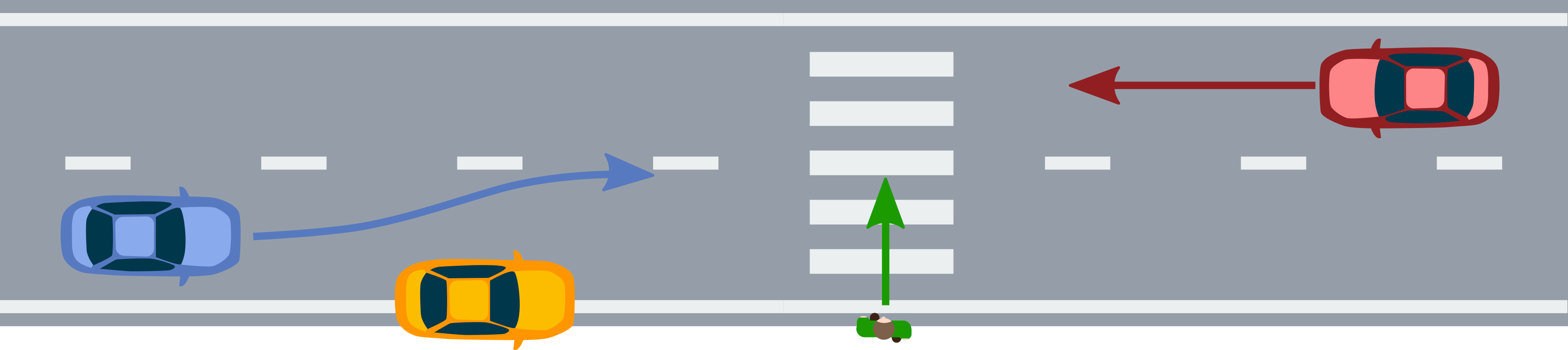}
	\caption{A typical urban scenario: Pedestrians crossing the street, a parked vehicle (yellow) blocking part of the lane and oncoming traffic (red). The ego vehicle is displayed in blue.}
	\label{fig:intro}
\end{figure}

A typical urban scene is presented in \refFigure{fig:intro}. The blue ego vehicle needs to overtake the stationary yellow vehicle and consider oncoming traffic and pedestrians crossing the street. Planning architectures that separate tactical maneuver selection and trajectory planning create handicaps in these types of situations. First of all, the separation may lead to sequences of high level actions that are physically not feasible. While this is typically handled by introducing additional safety margins, it limits the planner's ability to navigate in highly constrained environments with multiple obstacles. Second, if the tactical planner does not take the topology of the planning problem into account, the high-level sequence of actions passed to the trajectory planner may not be consistent with the past.

\subsection{Related Work}
\label{subsec:related_work}
Spatiotemporal motion planning approaches can be divided into path-velocity decomposition approaches \cite{Qian2016c}, sampling-based approaches \cite{Mcnaughton2011, Werling2010a} and optimization methods using Model Predictive Control \cite{Ziegler2014}.

Planning architectures which decouple the spatiotemporal problem into path- and speed-profile planning reduce the computational costs by a considerable amount. The decomposition into a high-level path planning problem and a trajectory planning problem works well for traffic circles or simple crossings when a predefined path does not change. However, path-velocity decomposition provides poor results for complex scenarios with moving obstacles.

Sampling-based methods are able to deal with non-convex constraints. \citet{Mcnaughton2011} present a spatial-temporal state-lattice-based approach based on dynamic programming. Due to the necessary state discretization with lattice-based methods, it is only suitable for highway driving.
\citet{Werling2010a} propose a state-lattice based implementation that generates sampled quintic polynomials. %
The jerk-optimal trajectories are first selected in local coordinates and then transformed to Cartesian coordinates to check for collisions. This computationally expensive transformation combined with the curse of dimensionality resulting from a high state space discretization limits the ability of this approach to reactively avoid obstacles.

Creating multiple trajectories in a receding horizon fashion introduces the problem of consistent trajectory selection over time. \citet{Gu2016} use a sampling-based trajectory planner to obtain trajectory candidates for the combinatorial motion planning problem. In order to avoid oscillation between multiple maneuvers, they group the generated trajectories by topological properties afterwards and impose consistency over time. \citet{Sontges2017} use a similar concept of topological grouping for the analysis of reachable sets. Each driving corridor then corresponds to a different high-level decision sequence. 

Local optimization methods formulating the motion problem as an optimal control problem do not suffer from any discretization errors in contrast to sampling based methods. \citet{Ziegler2014} present a spatiotemporal non-linear local optimization scheme. Due to the non-linear model formulation, computation time highly depends on the quality of the initialization. As this approach only guarantees to find local optima, it requires a preprosessing layer decomposing the combinatorial space to set up collision constraints for each maneuver variant \cite{Bender2015}. However, a generic constraint generation for complex scenarios still poses a major problem to the decomposition of the state space. 

In order to deal with the combinatorial aspects, \citet{Zhan2017} introduce a planning framework that plans longitudinal and lateral spatial movements separately to reduce computational costs. They use a layered graph-search approach and combine lateral and longitudinal motion using quadratic optimization. They classify environmental objects into point-overlap, line-overlap and undecided-overlap to deal with the combinatorial aspect. However, their search-based approach introduces longitudinally discretized actions that may not cover the optimal solution. 

\subsection{Contribution of this Paper}
In this work, we propose a planning approach which plans trajectories for multiple maneuver types. 
We adapt the idea from \cite{Zhan2017} of a generic environmental representation of obstacles along the ego vehicle's reference curve but apply it to optimization based planning. This way, we acknowledge the combinatorial aspect of motion planning and reduce the number of maneuvers passed to the trajectory planner. We generate maneuver envelopes, that represent constraints to fully characterize a local trajectory optimization problem. Based on the optimization programs proposed in \cite{Gutjahr2016, Gutjahr2016a}, we separate longitudinal and lateral motion planning to reduce the computational costs. We calculate optimal trajectories in local coordinates for multiple maneuver types. We then need to select the best maneuver based on motion optimality and maneuver consistency, which is why we group the planned trajectories to maneuver variants. We will use the idea of topological trajectory grouping from \cite{Gu2016} and apply it to an optimization-based trajectory planner to allow for a reasoning about the planned maneuver variant.

To summarize, we contribute
\begin{itemize}
	\item a novel approach for a fused trajectory planning and maneuver selection, solving the existing problem of feasibility of pre-defined maneuvers,
	\item an optimization-based framework able to deal with combinatorial options and
	\item the demonstration of the technical abilities in a challenging scenario.
\end{itemize}

This work is further organized as follows: \Secref{sec:problem} defines the problem this paper aims to solve. The proposed method is presented in \Secref{sec:method}. \Secref{sec:evaluation} evaluates the algorithm's abilities followed by a discussion in \Secref{sec:conclusion}.

%% file: problem.tex
\section{Problem Statement and Definitions}
\label{sec:problem}

Given a reference curve as a series of points with no requirements on smoothness, we aim to find a collision-free and comfortable trajectory, roughly following the reference curve. The motion planner must account for static and dynamic obstacles. A superior strategy module responsible for high-level decisions like lane changes passes both a reference curve and a reference velocity $v_{ref}$ to the trajectory planning module. This way, map-based information such as recommended velocities while turning can be incorporated in the planning module. The trajectory planner should output a sequence of states in world coordinates that can be passed to a trajectory tracking controller. This work omits the uncertainty about the state of other traffic participants.

\begin{figure}[t]
\vspace{0.2cm} %
  \centering 
    \includegraphics[width=6.5cm]{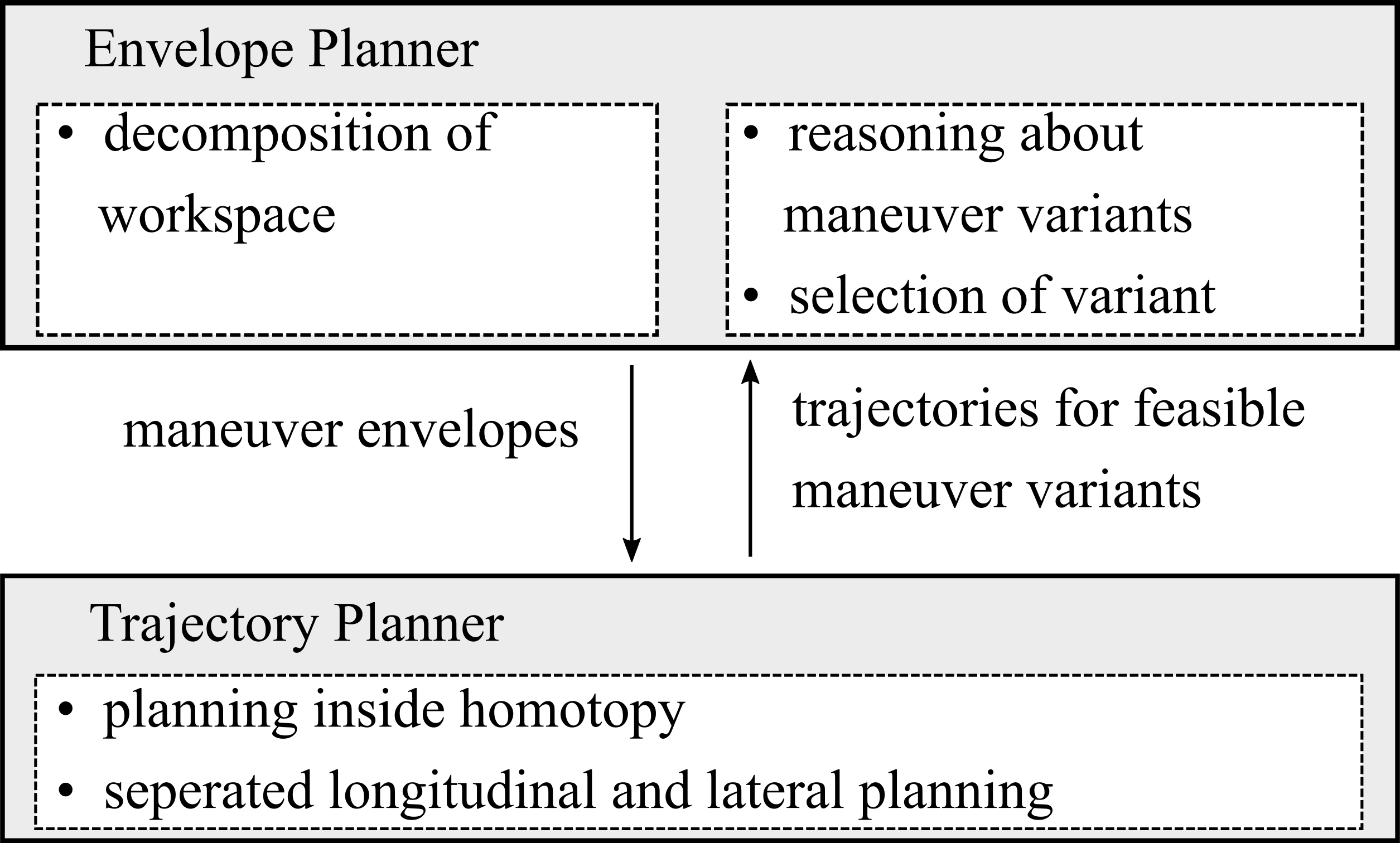} 
  \caption{Architectural overview of the proposed method consisting of an envelope planner passing homotopic planning candidates to a trajectory planner. The solved trajectories are passed back and evaluated for optimality and semantic consistency.} 
  \label{fig:achitecture} 
\end{figure}

\refFigure{fig:achitecture} shows an architectural overview of our approach. An envelope planner decomposes the spatiotemporal planning problem into multiple sub-problems, which we will call \textit{maneuver envelopes}. These maneuver envelopes are then passed to the trajectory planner. Each envelope leads to a local optimization problem, for which a set of \textit{homotopic trajectories} exists. Homotopic trajectories are co-terminal trajectories that can be continuously deformed from one to another without intersecting an obstacle \cite{Kim2012}. We will use these maneuver envelopes to impose linear collision avoidance constraints to the local optimization problem instead of relying on a suitable initialization.

However, the maneuver envelopes do not contain the temporal passing order of the objects $O_i$, which motivates us to adapt the definition of a \textit{maneuver variant} from \cite{Bender2015} to be a set of homotopic trajectories.
Following the ideas of \cite{Gu2016}, we distinguish different maneuver variants using topological distinction (\textit{How does the trajectory avoid obstacles?}) and sequential distinction (\textit{What overtaking order does it follow?}). \citet{Sontges2017} argue that obeying the mathematical definition of homotopy does not lead to a grouping of trajectories suitable for autonomous driving. In order to semantically describe the planned trajectories for reasoning in \ref{sec_sub:optimal_maneuver_selection}, we construct our semantic language $\mathrm{L}$ as following:

\begin{equation}
\mathrm{L} := \mathrm{L} \: \texttt{then} \: \mathrm{L} \:\parallel \:\mathcal{O} \: \texttt{is passed} \: \mathcal{H} \: \parallel \: \mathrm{L} \: \texttt{and} \: \mathrm{L}
\end{equation}
with 
\begin{equation} 
\mathcal{O} := \big\{O_i\big\}
\end{equation}
and
\begin{equation} 
\mathcal{H} := \texttt{left} \parallel \texttt{right}
\end{equation}

We state events to happen sequentially ($\texttt{then}$) or simultaneously ($\texttt{and}$).

%% file: method.tex
\section{Planning and Reasoning Framework}
\label{sec:method}

\subsection{Combinatorial Decomposition}
\label{sec_sub:combinatorial_decomposition}
We introduce an environment representation and decompose the combination problem into convex sub-problems. We define longitudinal and lateral rules for formulating convex state constraints for each obstacle type. This set of rules essentially allows us to reduce the number of sub-problems.

\subsection*{Spatiotemporal Environmental Representation}
Based on the predicted motion of the obstacle in relation to the reference curve of the ego vehicle, we construct convex hulls around the obstacle. We then derive free space-time decision envelopes $\bm{\zeta}^{o,\Delta}_{t_p}$ for each obstacle $o$ and decision $\Delta$ in a local reference system as 

\begin{equation} 
\bm{\zeta}^{o,\Delta}_{t_p} = 
\begin{bmatrix} 
\bm{\zeta}^{o,\Delta}_{long, max}\\[3pt]
\bm{\zeta}^{o,\Delta}_{long, min}\\[3pt]
\bm{\zeta}^{o,\Delta}_{lat, max}\\[3pt]
\bm{\zeta}^{o,\Delta}_{lat, min}\\[3pt]
\end{bmatrix} = 
\begin{bmatrix} 
s_{max}\\[3pt]
s_{min}\\[3pt]
d_{left}(s_i)\\[3pt]
d_{right}(s_i)\\[3pt]
\end{bmatrix}
\in R^{2+2n_s}
\end{equation}
at each prediction time step $t_p$ with the arc length $s$ and the perpendicular offset $d$. $n_s$ denotes the number of spatial sampling points $s_i$. This formulation allows us to represent obstacle-related longitudinal and lateral constraints in local coordinates independently of the future ego motion of the vehicle.

The longitudinal constraints are derived by calculating the spatial length at which the projected occupied space of the obstacle overlaps with the free-space of the ego vehicle along the reference line. We approximate this free-space with a polygon with the width of the ego vehicle's current lane. The lateral constrains are derived based on a distance calculation between the reference line and the obstacle at each spatial support point $s_i$.

\begin{figure}[t]
\vspace{0.2cm} %
\centering
\subfigure[]{\label{subfig:combinatorial_options_no-overlap:a}\includegraphics[width=0.48\columnwidth]{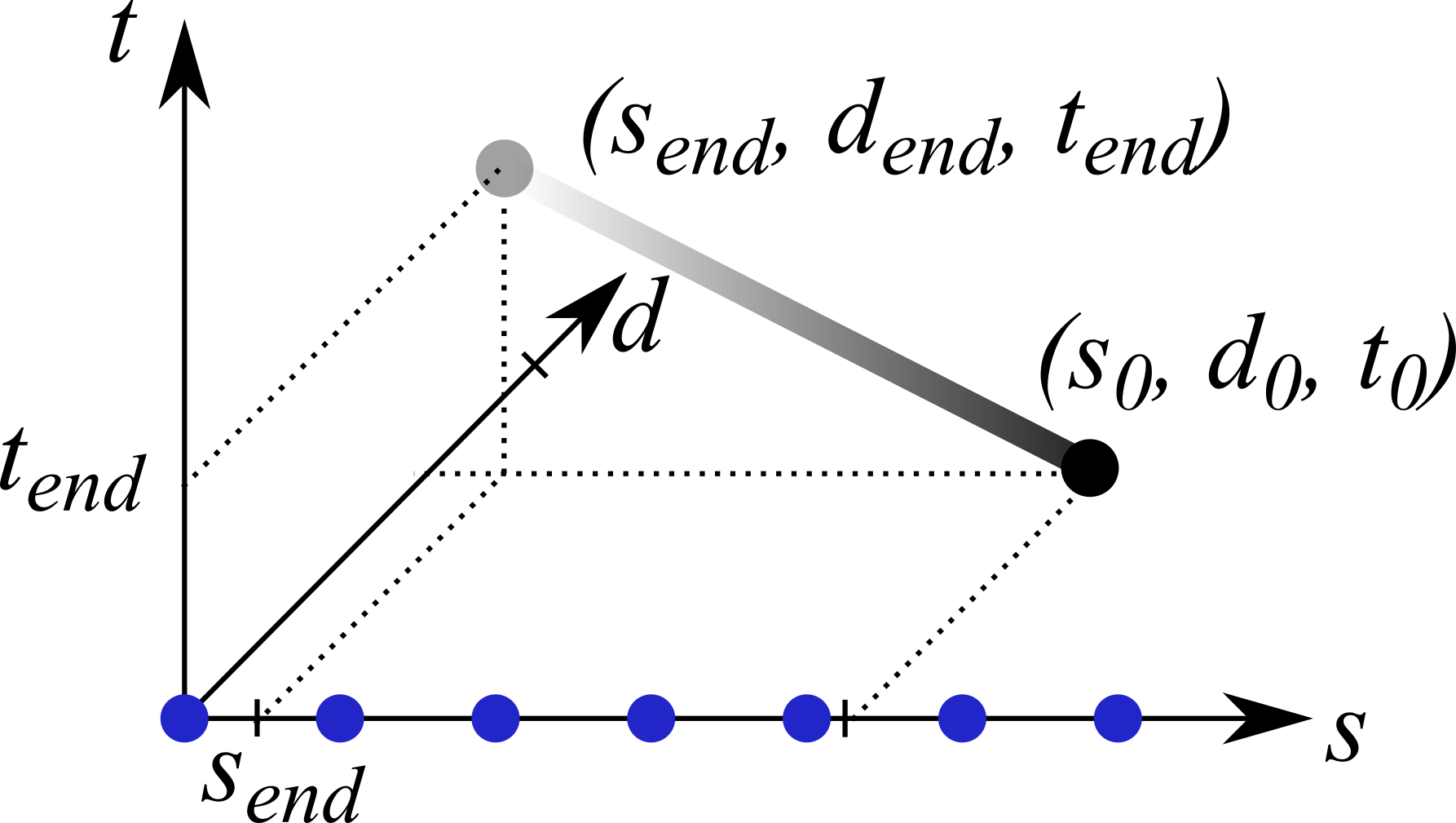}}\hfill
\subfigure[]{\label{subfig:combinatorial_options_no-overlap:b}\includegraphics[width=0.38\columnwidth]{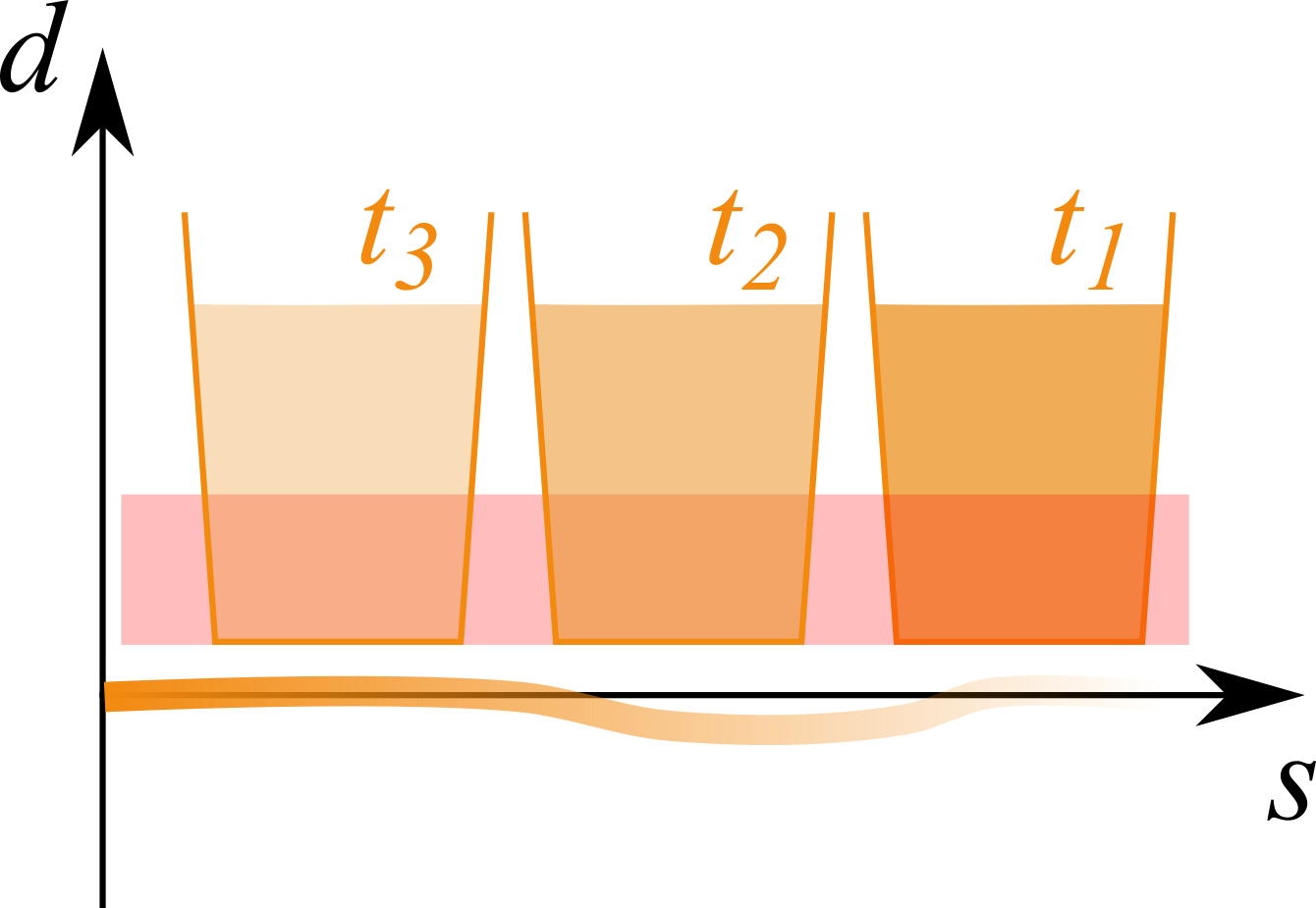}}
\caption{\subref{subfig:combinatorial_options_no-overlap:a} illustrates the motion of a non-overlapping obstacle in a local reference frame along the reference curve with the arc length $s$ and the lateral offset $d$. Blue reference points represent the ego vehicle's reference curve. As there is no overlap with the reference curve, the obstacle only needs to be considered laterally. \subref{subfig:combinatorial_options_no-overlap:b} shows the resulting lateral constraints (orange). The projected occupied space of the obstacle is shown in red. A possible trajectory of the ego vehicle is color-coded in orange indicating temporal progress.}
\label{fig:combinatorial_options_no-overlap}
\end{figure}

\begin{figure}[ht]
\centering
\subfigure[]{\label{subfig:combinatorial_options_line-overlap:a}\includegraphics[width=0.48\columnwidth]{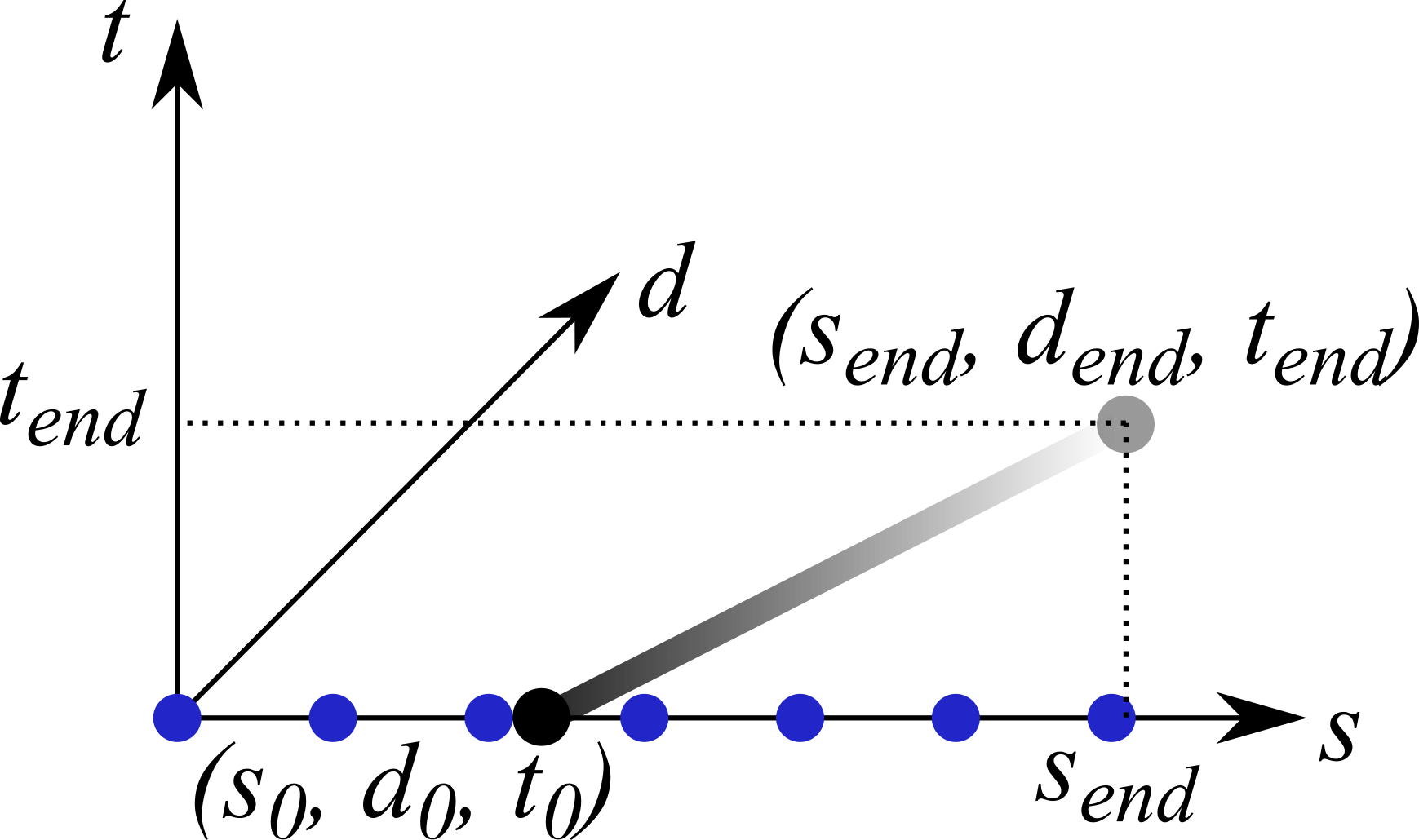}}\hfill
\subfigure[]{\label{subfig:combinatorial_options_line-overlap:b}\includegraphics[width=0.38\columnwidth]{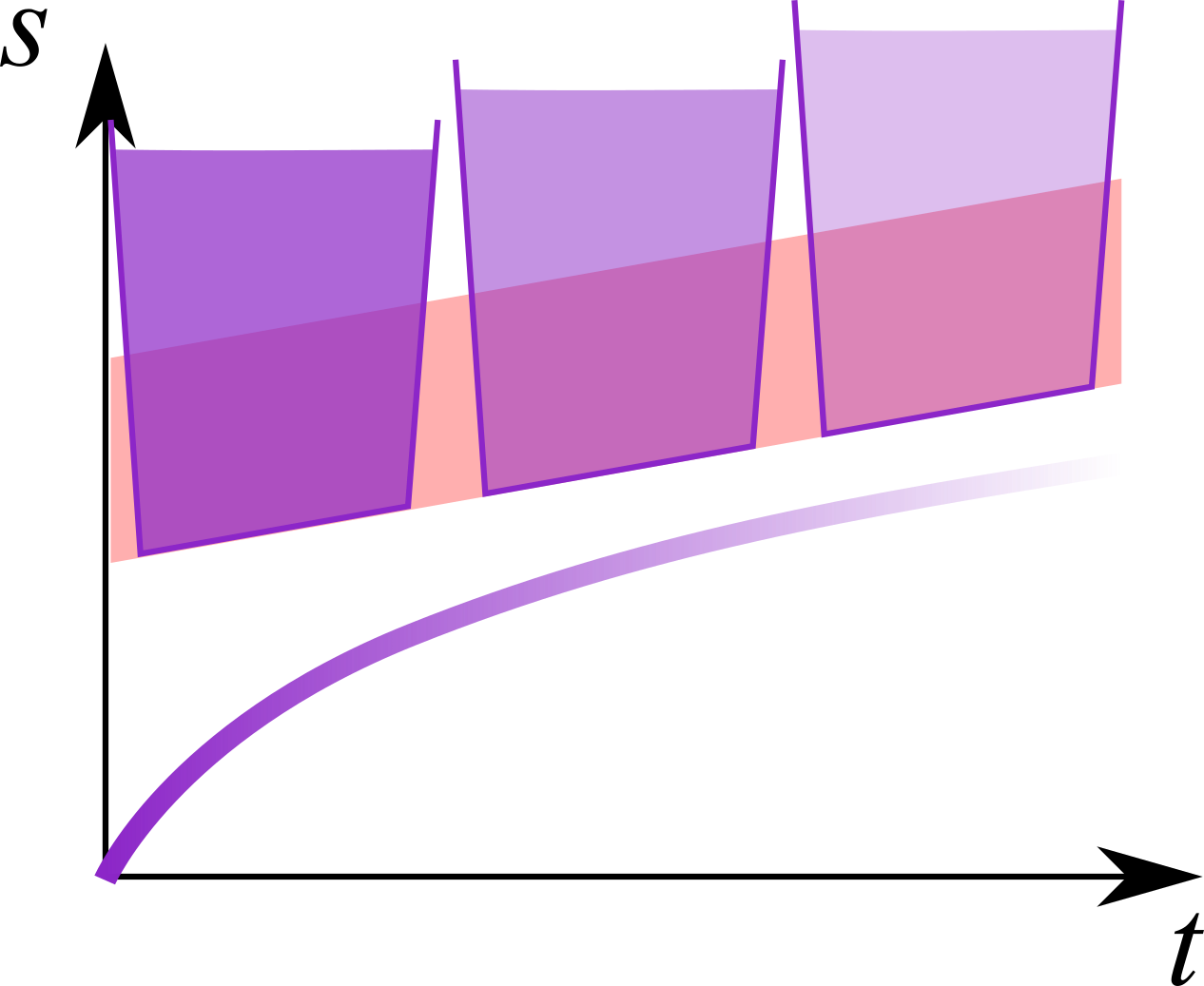}}
\caption{\subref{subfig:combinatorial_options_line-overlap:a} shows the motion of a line-overlapping obstacle. As displayed in \subref{subfig:combinatorial_options_line-overlap:b}, the obstacle type needs to be accounted for longitudinally, imposing state constraints on the spatial coordinate $s$.}
\label{fig:combinatorial_options_line-overlap}
\end{figure}

\begin{figure}[tb]
\vspace{0.2cm} %
\centering
\subfigure[]{\label{subfig:combinatorial_options_point-overlap:a}\includegraphics[width=0.48\columnwidth]{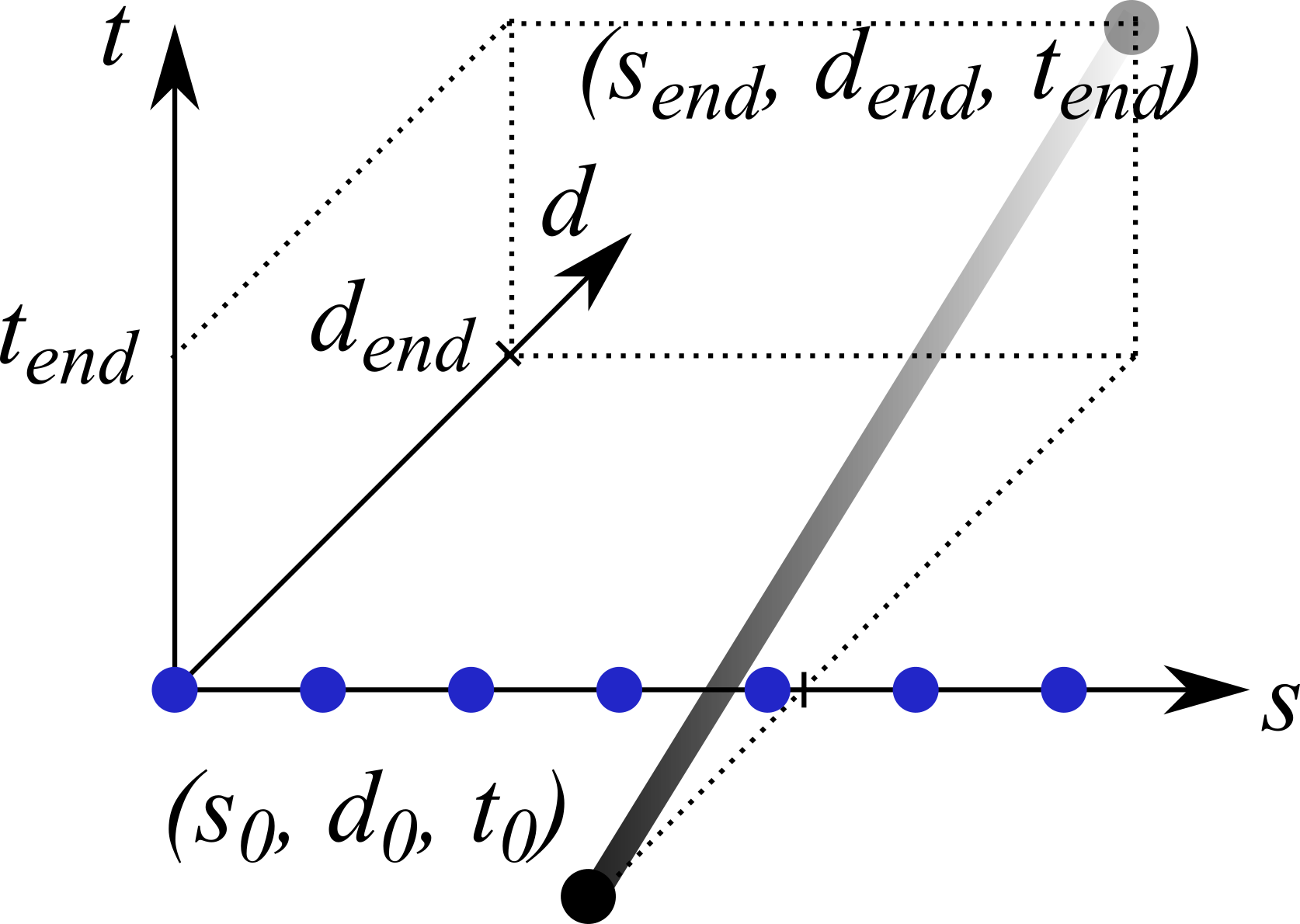}}\hfill
\subfigure[]{\label{subfig:combinatorial_options_point-overlap:b}\includegraphics[width=0.4\columnwidth]{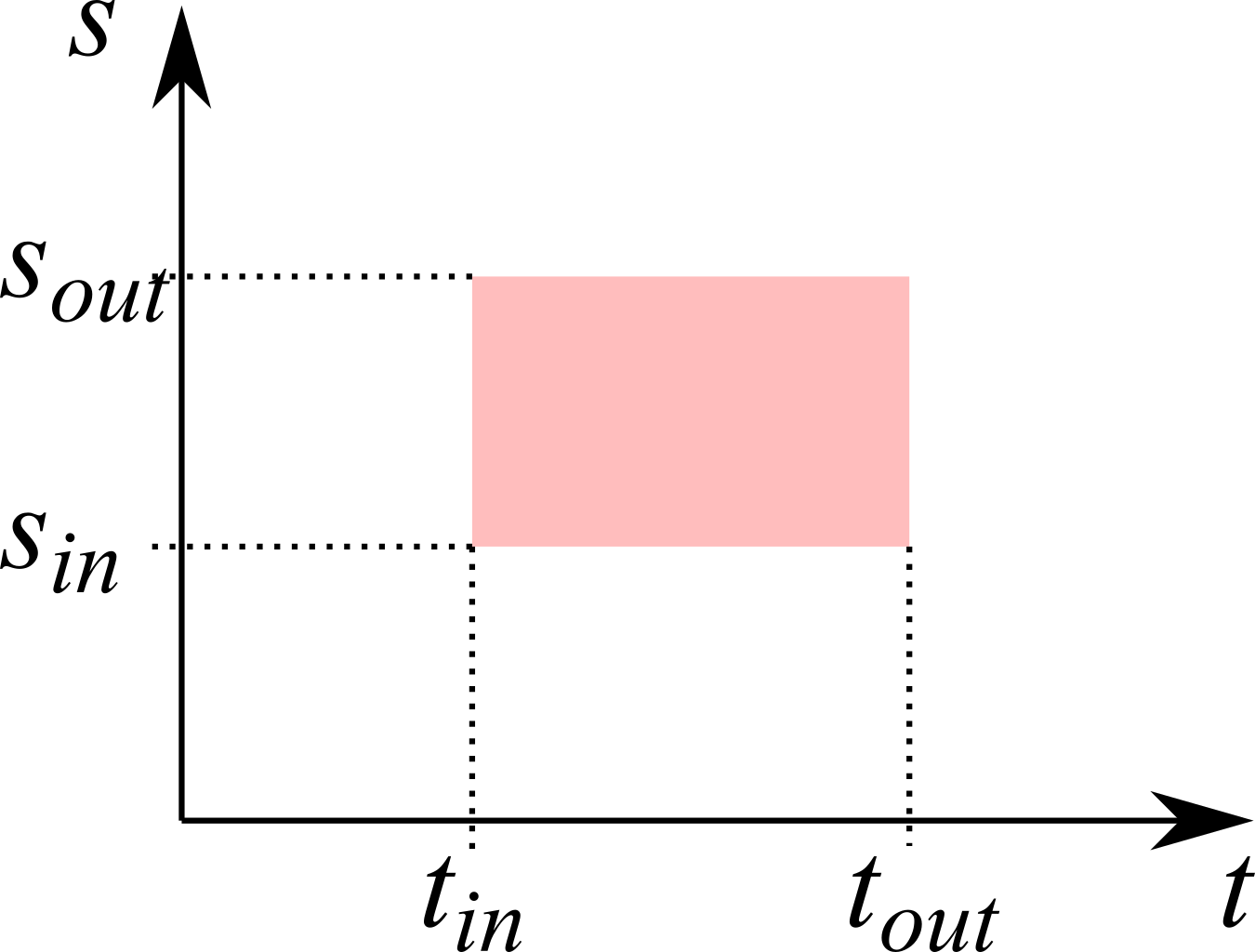}}
\subfigure[]{\label{subfig:combinatorial_options_point-overlap:c}\includegraphics[width=0.4\columnwidth]{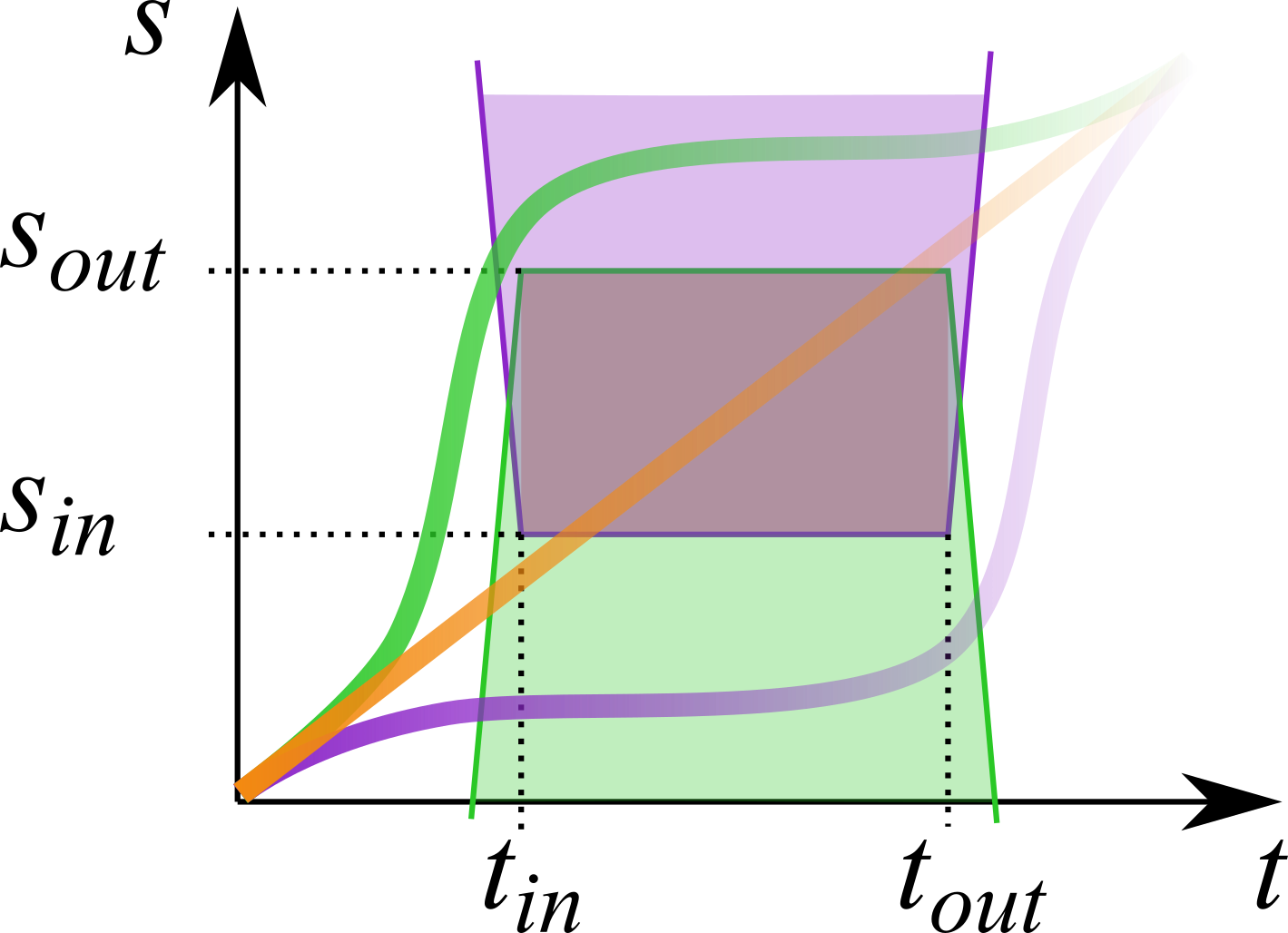}}\hfill
\subfigure[]{\label{subfig:combinatorial_options_point-overlap:d}\includegraphics[width=0.4\columnwidth]{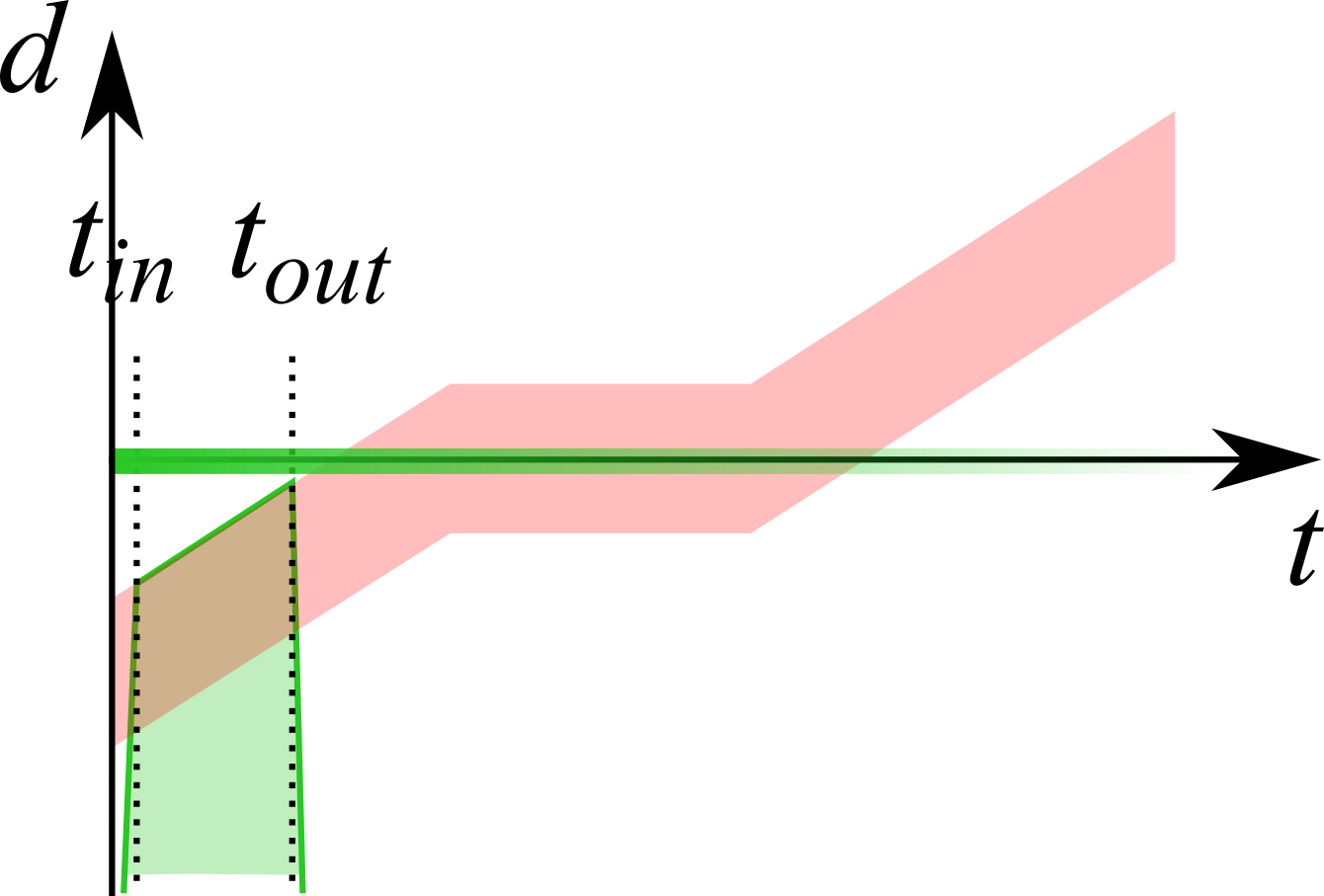}}
\subfigure[]{\label{subfig:combinatorial_options_point-overlap:e}\includegraphics[width=0.4\columnwidth]{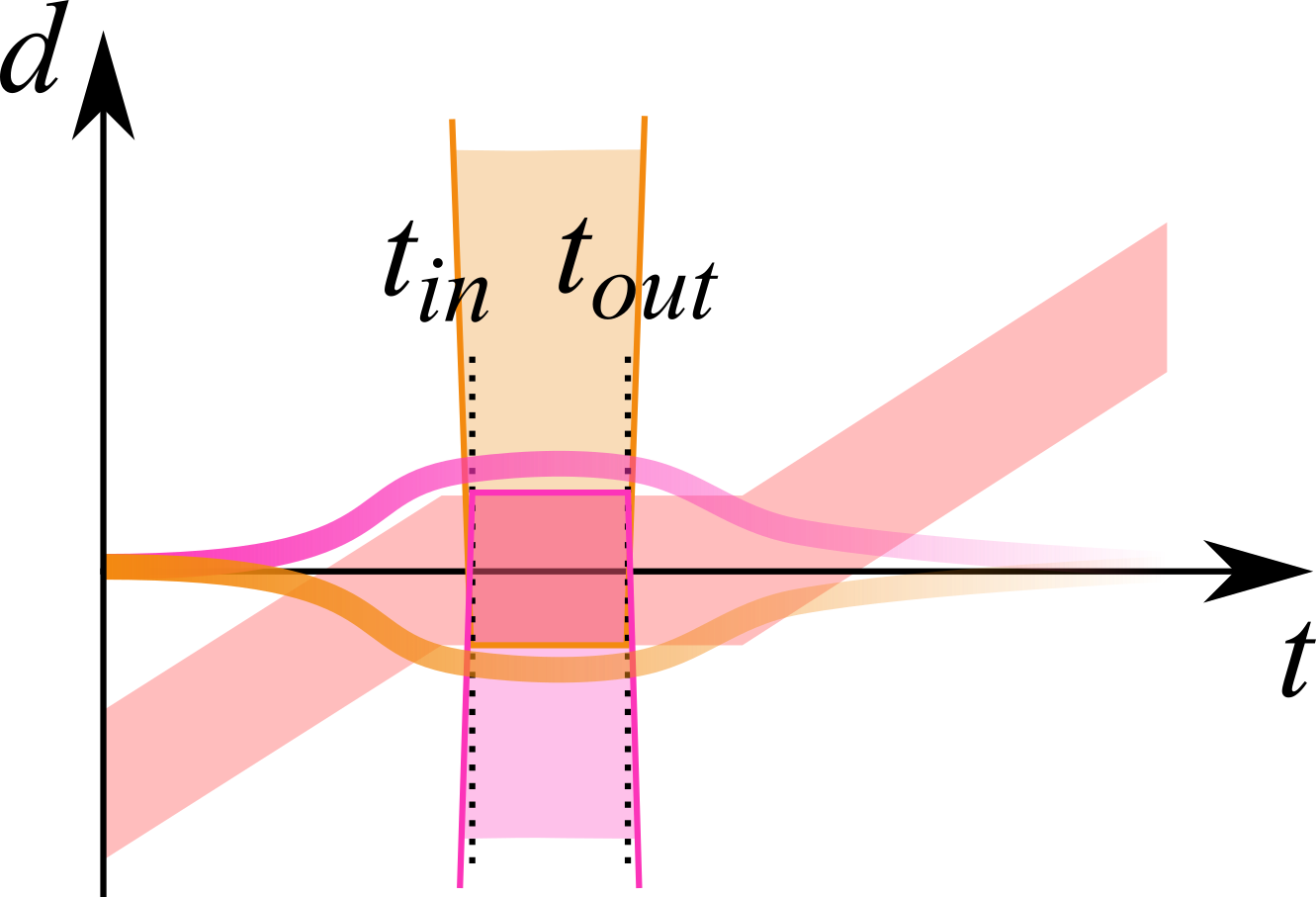}}\hfill
\subfigure[]{\label{subfig:combinatorial_options_point-overlap:f}\includegraphics[width=0.4\columnwidth]{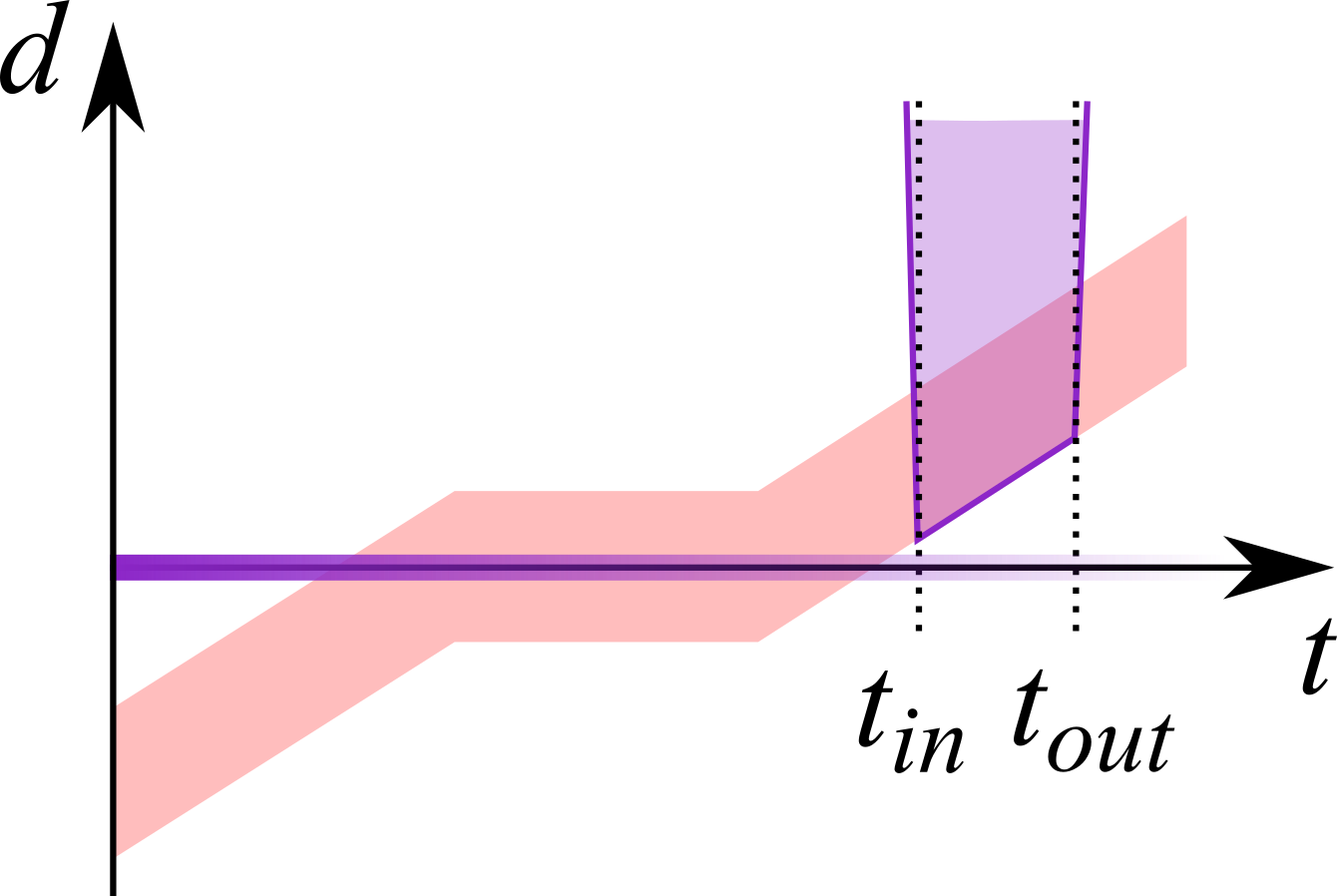}}
\caption{\subref{subfig:combinatorial_options_point-overlap:a} illustrates the motion of a point-overlapping obstacle. \subref{subfig:combinatorial_options_point-overlap:c} shows the combinatorial options passing the obstacle before (green), after (purple) or avoiding it left or right (orange). The respective lateral constraints for the maneuvers passing before \subref{subfig:combinatorial_options_point-overlap:d}, avoiding left or right (magenta and orange) in \subref{subfig:combinatorial_options_point-overlap:e}) and passing after \subref{subfig:combinatorial_options_point-overlap:f} are displayed.}
\label{fig:combinatorial_options_point-overlap}
\end{figure}

Inspired by \citet{Zhan2017}, we classify obstacles into \textit{non-overlapping} (\refFigure{subfig:combinatorial_options_no-overlap:a}), \textit{line-overlapping} (\refFigure{subfig:combinatorial_options_line-overlap:a}) and \textit{point-overlapping} (\refFigure{subfig:combinatorial_options_point-overlap:a}) obstacles. 
For \textit{non-overlapping} obstacles, there exists no intersection between the obstacle's predicted future motion and the reference path of the ego vehicle. \refFigure{subfig:combinatorial_options_no-overlap:a} shows an example where the obstacle is parallel to the reference path (e.g. oncoming traffic). We limit the set of tactical decisions for this class of obstacles by stating that the ego vehicle should only pass the obstacle on the side of the reference path (\refTable{tab:obstacle_choices}). Therefore, a decision on which side to avoid the obstacle is not needed. \refFigure{subfig:combinatorial_options_no-overlap:b} displays the resulting state constraints and a possible ego trajectory avoiding the obstacle.

\textit{Line-overlapping} obstacles are characterized through a line-wise overlapping to the vehicle's configuration space along the reference curve. This could for example be a preceding vehicle. \refFigure{subfig:combinatorial_options_line-overlap:b} shows the imposed state constraints by a preceding vehicle slowing down the ego vehicle.

\textit{Point-overlapping} obstacles have a fixed entry point and a fixed exit point from the configuration-space defined through the reference path. This could be vehicles at an intersection, pedestrians on a crosswalk or other agents intersecting with the reference path. \refFigure{fig:combinatorial_options_point-overlap} displays our constraint formulations for the combinatorial options for a \textit{point-overlapping} obstacle. We define four possible options: Passing the pedestrian before or after, or avoiding it on the left or right, see \refFigure{subfig:combinatorial_options_point-overlap:c}). For each of them, we derive longitudinal and lateral state constraints. As \refFigure{subfig:combinatorial_options_point-overlap:e} shows, when maintaining speed intending to avoid the obstacle, lateral constraints for left or right need to be considered. The lateral constraints for passing before and after are illustrated in \refFigure{subfig:combinatorial_options_point-overlap:d} and \ref{subfig:combinatorial_options_point-overlap:f}. 

The possible options for each obstacle class are summarized in \refTable{tab:obstacle_choices}.

\begin{table}[h]
\centering
\caption{Tactical decisions for passing or avoiding obstacles.}
\begin{tabular}{lcccc}
\hline
Obstacle Type & \multicolumn{4}{c}{Possible Tactical Decisions $\Delta_j$} \\
\cline{2-5}
          & $\Delta_1$ & $\Delta_2$ & $\Delta_3$ & $\Delta_4$ \\
\hline
Point-overlapping & before & after & right & left\\
Line-overlapping & & after & right & left\\
Non-overlapping & & & right/left & \\
\hline
\end{tabular}
\label{tab:obstacle_choices}
\end{table}

\subsection*{Combinatorial Planning Scheme}
\label{sec_sub:combinatorial_scheme}

For each prediction time step $t_p$, we derive the maneuver envelopes $\bm{\zeta}^{m}_{t_p}$ by creating all combinations of the obstacle's free space-time decisions envelopes $\bm{\zeta}^{\Delta}_{t_p}$ and merging them according to
\begin{equation} 
\bm{\zeta}^{m}_{t_p} = 
\left \{
\begin{aligned}
& \min && \bm{\zeta}^{o, \Delta}_{long, max}\\ 
& \max && \bm{\zeta}^{o, \Delta}_{long, min}\\ 
& \min && \bm{\zeta}^{o, \Delta}_{lat, max}\\ 
& \max && \bm{\zeta}^{o, \Delta}_{lat, min}\\
\end{aligned} \right. \;\;\;\forall o_i, \Delta_j. 
\label{equation:maneuver_envelopes}
\end{equation}

\begin{figure}[t]
\vspace{0.2cm} %
  \centering 
    \includegraphics[width=\columnwidth]{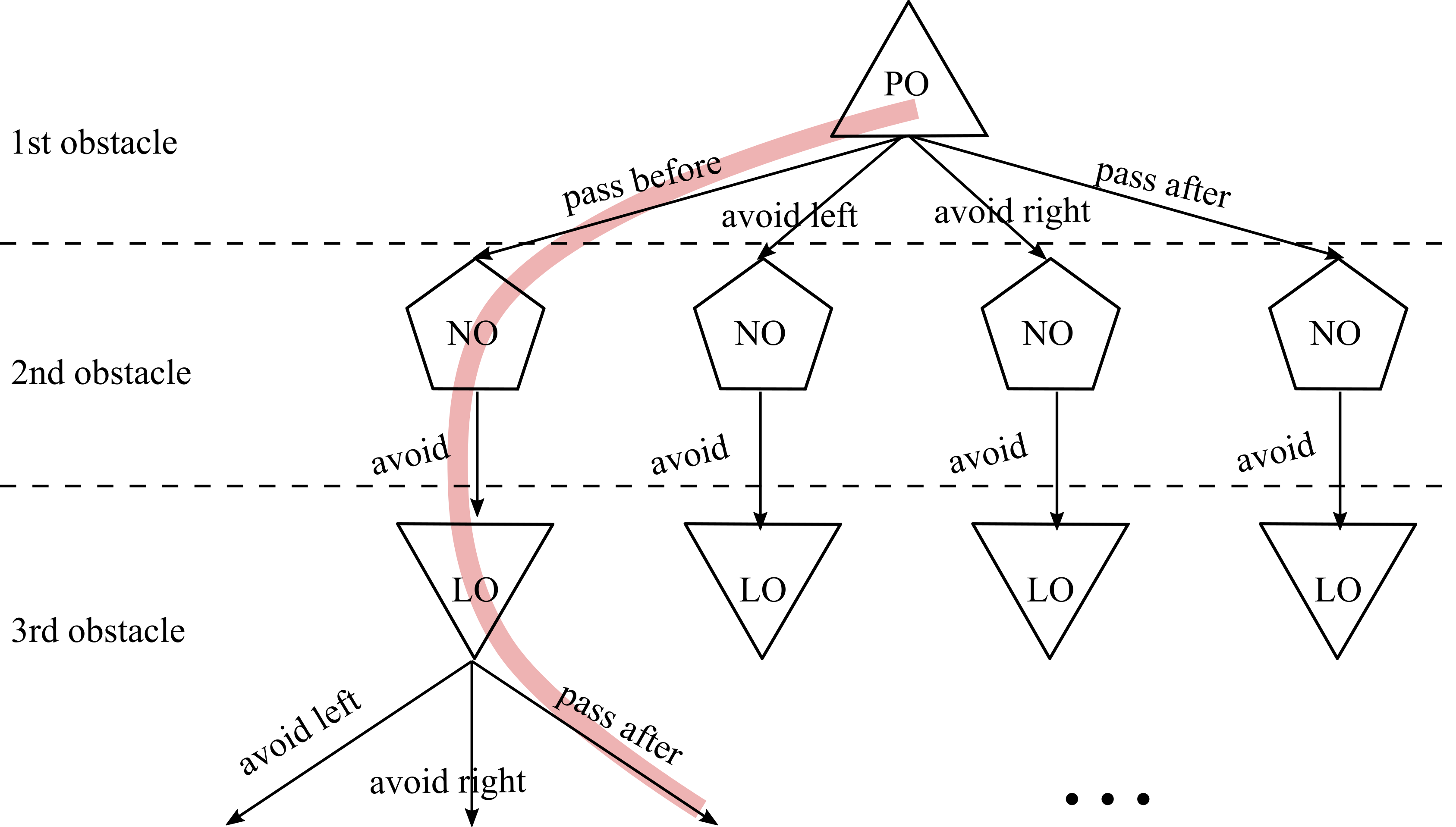} 
  \caption{Tactical decisions for a scene with three obstacles: a point-overlapping obstacle (1st), a non-overlapping obstacle (2nd) and a line-overlapping obstacle (3rd). The red line indicates a tactical decision sequence, that represents a maneuver envelope. The sequential ordering of the objects is not encoded here.}
  \label{fig:combinatorial_exploration} 
\end{figure}

\refFigure{fig:combinatorial_exploration} displays a tree with the tactical decisions for a set of a point-overlapping, a non-overlapping and a line-overlapping obstacle. Traversing the tree from the root to a leaf will be called \textit{maneuver sequence} $m$ for the remainder of this paper. Invalid maneuver envelopes are pruned, such as if $d_{left} < d_{right}$ with $d_{left} > 0$ and $d_{right} > 0$. As the number of sequences grows exponentially with the number of obstacles, we plan to reduce this set even further in the future by the use of heuristics.

The valid maneuver envelopes are passed to the trajectory planner. The constraints of each maneuver envelope form a convex state space, making sure the trajectory planner can only converge to one optimum. First, a longitudinal trajectory candidate is generated for each envelope. If the problem is infeasible, the formulation as a linear quadratic program allows us to quickly terminate the optimization if it does not converge. For all longitudinal candidates, we optimize the lateral behavior. If successful, we add the trajectory to the set of possible maneuvers. By reason of functional safety, we add a collision check in Cartesian coordinates for the projected motion to make sure that the coordinate transformations did not introduce any error that may lead to collisions. The optimal trajectory in regard to a set of criteria can then be selected from the set of possible trajectories (see \Secref{sec_sub:optimal_maneuver_selection}). The model predictive control is sped up by reusing the previous solution as an initialization.

\subsection{Trajectory Planning} 

\begin{figure}[t] 
\vspace{0.2cm} %
  \centering 
    \includegraphics[width=\columnwidth]{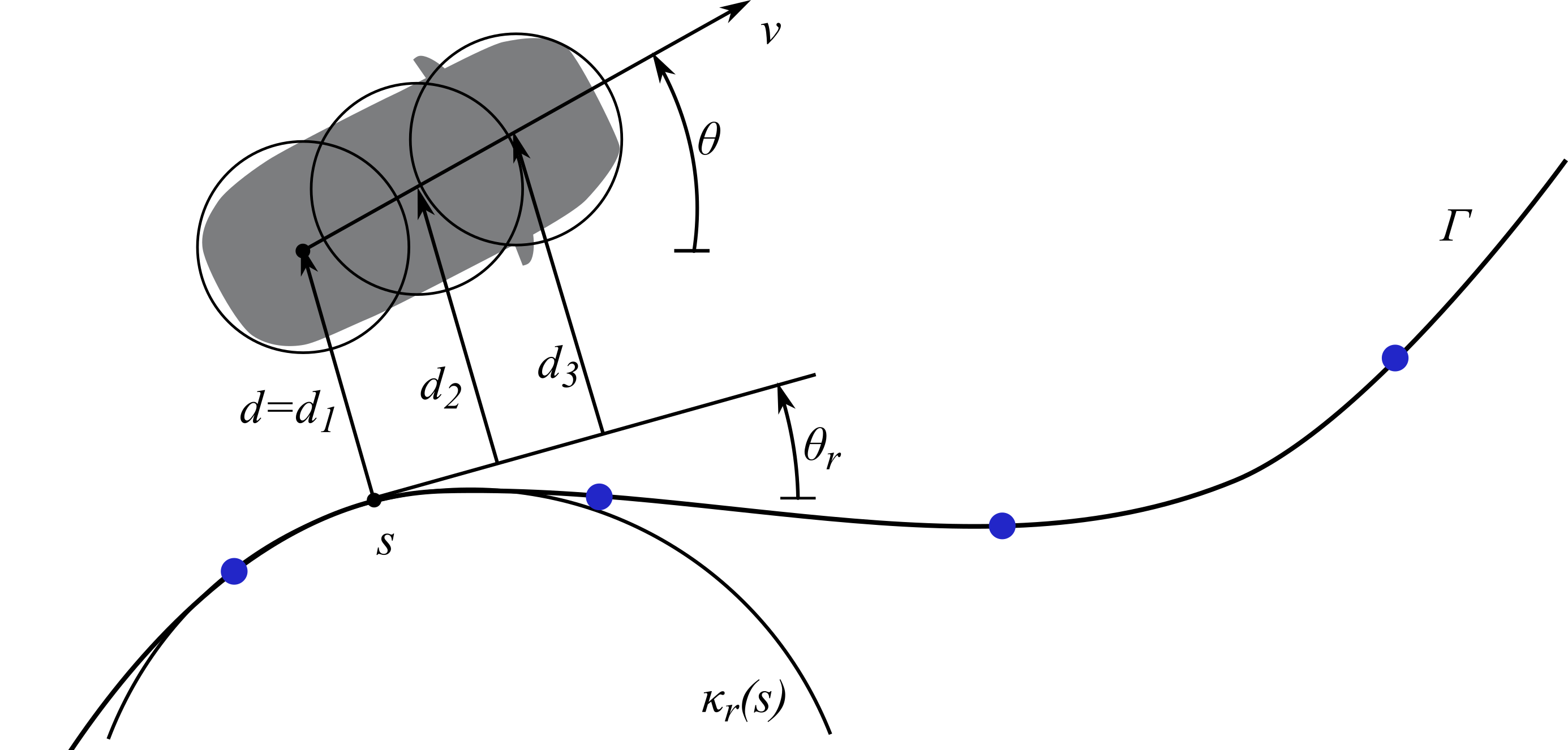} 
  \caption{Vehicle model along a reference curve $\mathit{\Gamma}$ in local street coordinates \cite{Gutjahr2016}. The blue waypoints indicate a discrete representation of the reference curve. The arc length $s$, the perpendicular offset $d$ to the reference curve and the orientation of the vehicle $\theta$ define the vehicle state. $d_1$, $d_2$, $d_3$ define the system output.} 
  \label{fig:frenet_frame} 
\end{figure}

The trajectory planning algorithm is running in a receding horizon fashion. In this section, we will mainly discuss a single optimization run representing a Model Predictive Control stage. In order to separate longitudinal and lateral motion, the trajectory optimization problem needs to be defined in a local reference frame as in \cite{Gao12, Gutjahr2016}, so-called \textit{Frenet coordinates}. Frenet coordinates define the motion along a reference curve $\mathit{\Gamma}(s)$ through the arc length $s$ and the lateral offset $d$ (see \refFigure{fig:frenet_frame}).

The decomposition into longitudinal and lateral motion allows us to handle the desired behavior separately, as e.g. the longitudinal motion may be subject to a high-level decision making entity. The spatial decomposition also simplifies the computational complexity of the motion planning scheme as it allows to formulate the trajectory planning problem through two linear sequential optimization programs. Additionally, the separated motion formulation allows to construct linear safety constraints for each motion.

In the following paragraph, we present the key ideas from the optimization schemes introduced in \cite{Gutjahr2016a, Gutjahr2016}. As the formulation uses a linearized model, no iterative scheme for approximating the non-linearity is needed, which significantly simplifies the problem and reduces the computational complexity. 

\subsection*{Optimal Control Problem} 
Both longitudinal and lateral optimization schemes introduced in \cite{Gutjahr2016a, Gutjahr2016} formulate constrained optimal control problems. With a linear model and a quadratic cost function, the control problem can be formulated over a prediction span $N$ and solved using the batch method \cite{Borrelli2017}. Given a state matrix $\bm{A}(t)$, an input matrix $\bm{B}(t)$, an error matrix $\bm{E}(t)$ and an output matrix $\bm{C}(t)$, the continuous system model  
\begin{equation} 
\bm{\dot{x}}(t) = \bm{A}(t) \bm{x}(t) + \bm{B}(t) \bm{u}(t) + \bm{E}(t) \bm{z}(t) 
\end{equation} 
\begin{equation} 
\bm{y}(t) = \bm{C}(t) \bm{x}(t)
\end{equation} 
can be discretized using Euler integration leading to 
\begin{equation} 
\label{equation:discrete_state_model} 
\bm{\hat{x}}(k+1) = \bm{\hat{A}}(k) \bm{x}(k) + \bm{\hat{B}}(k) \bm{\hat{u}}(k) + \bm{\hat{E}}(k) \bm{\hat{z}}(k) 
\end{equation} 
\begin{equation} 
\bm{\hat{y}}(k) = \bm{\hat{C}}(k) \bm{\hat{x}}(k).
\end{equation} 
By stacking together sequential states using the batch method, the quadratic program is transformed into a sequential quadratic program which can be solved using standard SQP solvers. With the discrete step size $k$ being represented by an index, the state vector sequence can be written as 
\begin{equation} 
\mathbf{x} = \begin{bmatrix} 
\bm{\hat{x}}^T_1, & \dots, & \bm{\hat{x}}^T_N \\ 
\end{bmatrix}. 
\end{equation} 
The state input $\mathbf{u}$, the state output $\mathbf{y}$ and the error $\mathbf{z}$ are defined analogously.  
The costs can then be formulated as 
\begin{equation} 
\label{equation:costs} 
J(\mathbf{x}, \mathbf{u}, \mathbf{x}_{ref}) = [\mathbf{x}-\mathbf{x}_{ref}]^T \mathcal{Q} [\mathbf{x}-\mathbf{x}_{ref}] + \mathbf{u}^T \mathcal{R} \mathbf{u} \,,
\end{equation} 
where $\mathcal{R}$ and $\mathcal{Q}$ denote cost weight matrices and $\mathbf{x}_{ref}$ denotes the reference state. 
Formulating states $x_j$ as outputs $y_j$ allows to express systems constraints as input constraints $u_{c,j}$.
\subsection*{Longitudinal Trajectory Planning} 
\label{sec_sub:longitudinal_planning} 
We use the linear time-variant system model presented by \citet{Gutjahr2016a}. The system model simplifies to  
\begin{equation} 
\bm{\dot{x}}(t) = \bm{A}(t) \bm{x}(t) + \bm{B}(t) \bm{u}(t) \,,
\end{equation} 
\begin{equation} 
\bm{y}(t) = \bm{C} \bm{x}(t) \,,
\end{equation}
where $\bm{x} = \begin{bmatrix} s, v, a, j\end{bmatrix}$ denotes the state and $\bm{u} = \begin{bmatrix} \ddot{a} \end{bmatrix}$ denotes the input. \citet{Gutjahr2016a} define the system output to constrain $s$ and $a$. We extend this to include the velocity $v$, as it allows us to prohibit backward motion. The system output is thus defined as $\bm{y} = \begin{bmatrix} s, v, a\end{bmatrix}$. The respective continuous state matrices can be found in the Appendix.
For the cost function \refEquation{equation:costs}, the cost weight matrices $\bm{\hat{Q}}(k)$ and $\bm{\hat{R}}(k)$ are defined as 
\begin{equation} 
\bm{\hat{Q}}(k) =  
\textrm{diag}(w_s(k), w_v(k), w_a(k), w_j(k))
\end{equation} 
and  
\begin{equation} 
\bm{\hat{R}}(k) =  
\begin{bmatrix} 
w_u(k)\\ 
\end{bmatrix}\,.
\end{equation} 
We end up solving the following longitudinal planning problem for each valid maneuver envelope $m$:
\begin{subequations}
\begin{align}
\forall m \quad \min_{u} \quad & J(\bm{x},\bm{u},\bm{x}_{ref})\\
s.t. \quad & \forall t \in [t_0, t_0+T] \\
& \bm{\dot{x}}(t) = f(\bm{x}(t), \bm{u}(t)) \\
& \bm{u}_{min} \leq \bm{u}(t) \leq \bm{u}_{max} \\
& \bm{y}^{m}_{min}(t) \leq \bm{y}(t) \leq \bm{y}^{m}_{max}(t)
\end{align}
\end{subequations}
with 
\begin{subequations}
\begin{align} 
\bm{y}^{m}_{max}(t) = &\begin{bmatrix} 
\bm{\zeta}^{m}_{long, max}(t), & v_{max}(t), & a_{max}(t) \\ 
\end{bmatrix},\\
\bm{y}^{m}_{min}(t) = &\begin{bmatrix} 
\bm{\zeta}^{m}_{long, min}(t), & v_{min}(t), & a_{min}(t) \\ 
\end{bmatrix}.
\end{align}
\end{subequations}
We express the output constraints as input constraints using batch matrices, cf. \cite{Gutjahr2016a}.
\subsection*{Lateral Trajectory Planning} 
\label{sec_sub:lateral_planning} 
We use the linear time-variant system model presented in \cite{Gutjahr2016}  
with the state $\bm{x} = \begin{bmatrix} d, \theta, \kappa, \theta_r, \kappa_r\end{bmatrix}$, the input $\bm{u} = \begin{bmatrix} \dot{\kappa} \end{bmatrix}$, the output $\bm{y} = \begin{bmatrix} d_1, d_2, d_3, \kappa\end{bmatrix}$ and the error $\bm{z} = \begin{bmatrix} \kappa_r \end{bmatrix}$.
The respective continuous state matrices can be found in the Appendix. The cost function minimizes the distance to the reference line $d$, the orientation difference between the vehicle and the reference line $\theta-\theta_r$, the curvature $\kappa$ and the change of curvature $\dot{\kappa}$. The usage of an explicit reference state $\bm{x}_{ref}$ is not necessary, as all state variables are desired to be zero. Formally, this leads to  
\begin{equation} 
\bm{Q}(k) =  
\begin{bmatrix} 
w_d(k) & 0 & 0 & 0 & 0 \\ 
0 & w_\theta(k) & 0 & -w_\theta(k) & 0 \\ 
0 & 0 & w_\kappa(k) & 0 & 0 \\ 
0 & -w_\theta(k) & 0 & w_\theta(k) & 0 \\ 
0 & 0 & 0 & 0 & 0 \\ 
\end{bmatrix} 
\end{equation} 
and  
\begin{equation} 
\bm{R}(k) =  
\begin{bmatrix} 
w_u(k)\\ 
\end{bmatrix}. 
\end{equation} 
The linearized model is only valid for small deviations $\theta - \theta_r$. The formulation can deal with reference curves even with high and discrete curvatures. This would allow the usage of a search-based path planner without a computationally costly post-optimization such as the one from \cite{Dolgov2010a}. The distance to obstacles is calculated based on the static reference curve. This means that the otherwise costly collision check is computed only once for each successful optimization instead of for each iteration step. 
Finally, we solve the following lateral optimization problem for all valid maneuvers $m$, for which the longitudinal planning succeeded.
\begin{subequations}
\begin{align}
\forall m \quad \min_{u} \quad & J(\bm{x},\bm{u})\\
s.t. \quad & \forall t \in [t_0, t_0+T] \\
& \bm{\dot{x}}(t) = f(\bm{x}(t), \bm{u}(t)) \\
& \bm{u}_{min} \leq \bm{u}(t) \leq \bm{u}_{max} \\
& \bm{y}^{m}_{min}(t) \leq \bm{y}(t) \leq \bm{y}^{m}_{max}(t)
\end{align}
\end{subequations}
with
\begin{subequations}
\begin{align}
\bm{y}^{m}_{max} =& \begin{bmatrix}
\bm{\tilde{\zeta}}^{m}_{lat, max}(t), \bm{\tilde{\zeta}}^{m}_{lat, max}(t), \bm{\tilde{\zeta}}^{m}_{lat, max}(t), \kappa_{max}(t) \\ 
\end{bmatrix}, \\
\bm{y}^{m}_{min} =& \begin{bmatrix} 
\bm{\tilde{\zeta}}^{m}_{lat, min}(t), \bm{\tilde{\zeta}}^{m}_{lat, min}(t), \bm{\tilde{\zeta}}^{m}_{lat, min}(t), \kappa_{min}(t) \\ 
\end{bmatrix}.
\end{align}
\end{subequations}
We make use of the batch approach again to express output as input constraints, cf. \cite{Gutjahr2016}. From the longitudinal planning, we know the longitudinal motion $s(t)$, which we use to transform the spatial-dependent $\bm{\zeta}^{m}_{lat}$ to time-dependent $\bm{\tilde{\zeta}}^{m}_{lat}$.
Similar to \cite{Gutjahr2016}, we use slack variables to relax the constraints for $d_{1,2,3}$ for the first three optimization support points, as the model error leads to infeasibilities close to obstacles when being used in a receding horizon fashion. 

\subsection{Reasoning and Maneuver Variant Selection}
\label{sec_sub:optimal_maneuver_selection}
After the construction of homotopic maneuver envelopes and their trajectory optimization, we need to select the best trajectory. 
For this, we first define our trajectory costs as
\begin{equation}
	 J_{traj} = \frac{1}{n}\sum_{i=1}^{n}{\Big[w_{r,a} a_i^2} + w_{r,j} j_i^2 + w_{r,d} \big(d_i-d_{d,i}\big)^2 +w_{r,\dot{\kappa}} \dot{\kappa_i}^2\Big].
\end{equation}
We incorporate comfort measures with the curvature derivative $\dot{\kappa_i}$, the longitudinal acceleration $a_i$ and the jerk $j_i$. We also integrate the proximity to other obstacles $d_i-d_{ref,i}$ as a safety measure into the cost functional. $d_{ref,i}$ represents the maximum possible distance to the left and right side
\begin{equation}
d_{ref,i} = \frac{1}{2}\Big(d_{left,i} + d_{right,i}\Big).
\end{equation}

Despite using an optimization-based planning approach, the temporal consistency of the selected maneuver will not necessarily hold, which potentially may lead to oscillating behavior. First of all, the receding horizon concept and the uncertainty of the environment provide new information to the trajectory planner at every planning stage. Second, the solutions obtained from the trajectory planner are only suboptimal as we simplify the trajectory planning to two separate planning tasks. This motivates to include a consistency cost term $J_{cons}$ which penalizes switching between maneuver sequences.
To quantify the costs, we need to describe the semantic consistency of the maneuver to the one previously selected. The semantic description needs to contain both topological as well as sequential information. As the maneuver envelopes from \Secref{sec_sub:combinatorial_decomposition} do not contain sequential information, we refer to the semantic language which we defined in \Secref{sec:problem} to map our obtained trajectories back to a maneuver variant. Similarly to \cite{Sontges2017}, we construct two-dimensional surfaces $U_i$ along $s$ and $t$ in the three-dimensional domain for each obstacle $O_i$. The intersection points between the ego trajectory $\tau$ and $U_i$ yield the sequential information $t_i$. The signed distance to the obstacle at $t_i$ yields the topological information.
This can be seen as an abstract function mapping possibly many trajectories onto one maneuver variant.

With these information, the semantic description for a maneuver variant can be automatically extracted from a given trajectory. 
It allows us to reason about the related maneuver variants and calculate the consistency costs
\begin{equation} 
J_{cons} = w_{r,c} \Big( N_{max} - \sum_{j=1}^M\, \delta (L_{j,t_i} \neq L_{j,t_{i-1}})\Big),
\end{equation}
where $N_{max}$ denotes the maximum number of elements in the language sequences at time $t_i$, $M$ the number of similar language items between the description $L_{t_i,j}$ at the current time step and the previous time step $L_{t_{i-1},j}$. For each similar item $L_{j}$ in shared order, $\delta(\cdot)$ outputs 1, otherwise 0.   This semantic reasoning adds an additional safety layer to our selection process. In the future, it could be enhanced by adding uncertainty information from the object detection module or incorporating occlusion.

The total costs $J_{total}$ for the selection process are then defined by

\begin{equation} 
J_{total} = J_{traj} + J_{cons} \,.
\end{equation}

%% file: evaluation.tex
\section{Evaluation}
\label{sec:evaluation}
We apply the proposed method to an urban driving scenario similar to the motivating example of this paper (see \refFigure{fig:intro}) with a crossing pedestrian $O_1$ at $v=0.5\si{\meter\per\second}$ (green), a static obstacle $O_2$ (yellow) and oncoming traffic $O_3$ at $v=10\si{\meter\per\second}$ (red). The chosen parameters are displayed in \refTable{tab:parameters_trajectory_optimization}, \refTable{tab:parameters_reasoning} and \refTable{tab:parameters_horizons}.

\begin{table}
\vspace{0.2cm} %
\centering
\caption{Weights for trajectory optimization}
\begin{tabular}{|l|l|l|l|l|l|l|l|l|}
\hline
$w_s$ & $w_v$ & $w_a$ & $w_j$ & $w_{\dot{j}}$ & $w_d$ & $w_{\theta}$ & $w_{\kappa}$ & $w_{\dot{\kappa}}$\\
\hline
\num{0} & \num{e3} & \num{10} & \num{e2} & \num{e3} & \num{e2} & \num{10} & \num{10} & \num{e3} \\
\hline
\end{tabular}
\label{tab:parameters_trajectory_optimization}
\end{table}

\begin{table}
\centering
\caption{Weights for reasoning}
\begin{tabular}{|l|l|l|l|l|}
\hline
$w_{r,a}$ & $w_{r,j}$ & $w_{r,d}$ & $w_{r,\dot{\kappa}}$ & $w_{r,c}$\\
\hline
\num{10} & \num{10} & \num{e2} & \num{e3} & \num{30} \\
\hline
\end{tabular}
\label{tab:parameters_reasoning}
\end{table}

\begin{table}
\centering
\caption{Parameters for planning and prediction horizons}
\begin{tabular}{l l l}
\hline
Horizon & $N$ & $dt [\si{\second}]$\\
\hline
Longitudinal Trajectory Optimization & $20$ & $0.2\si{\second}$\\
Lateral Trajectory Optimization & $20$ & $0.2\si{\second}$\\
Prediction & $10$ & $0.4\si{\second}$\\
\hline
\end{tabular}
\label{tab:parameters_horizons}
\end{table}

\begin{figure}[ht]
\centering
\subfigure{\label{subfig:a}\includegraphics[width=0.9\columnwidth, frame]{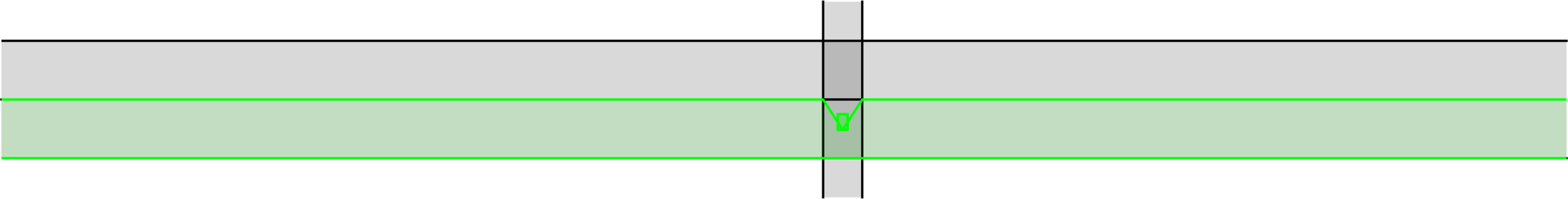}}
\subfigure{\label{subfig:a}\includegraphics[width=0.9\columnwidth, frame]{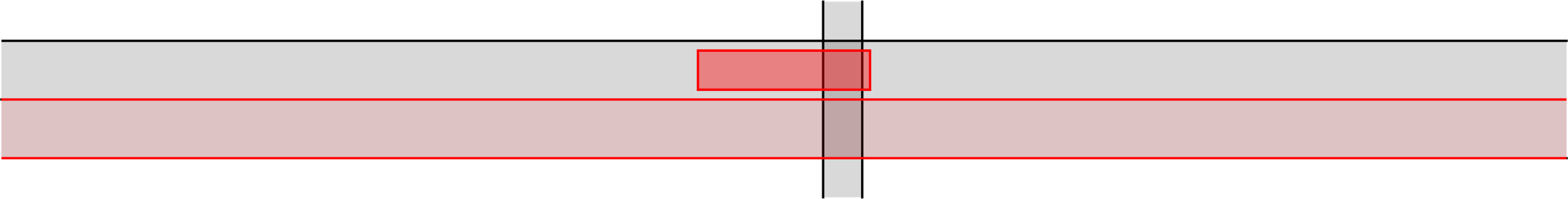}}
\subfigure{\label{subfig:b}\includegraphics[width=0.9\columnwidth, frame]{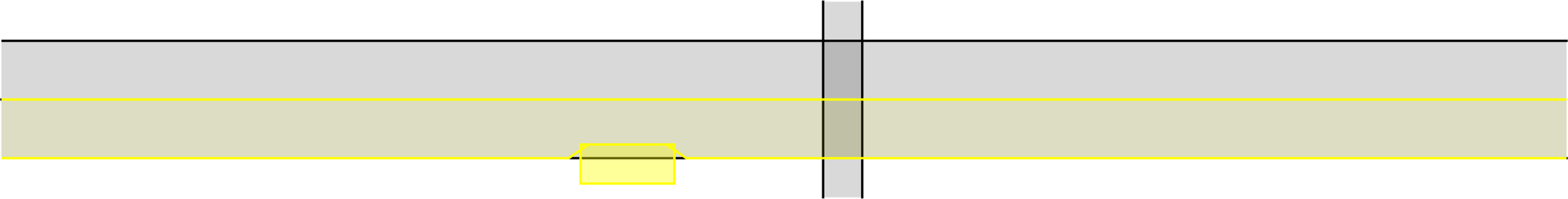}}
\subfigure{\label{subfig:b}\includegraphics[width=0.9\columnwidth, frame]{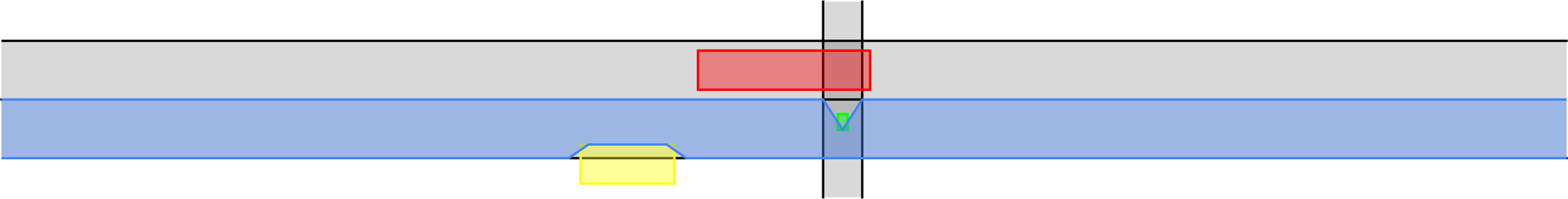}}
\caption{Free space-time decision envelopes $\bm{\zeta}^{o,\Delta}_{t_{p6}|t_0}$ for each obstacle at $t_0 =0\si{\second}$ for the prediction step $t_{p6}=2.4\si{\second}$ for the maneuver envelope $m_{15}$. The merged free space-time maneuver envelope $\bm{\zeta}^{m_{15}}_{t_{p6}|t_0}$ is displayed below.}
\label{fig:homotopic_constraints}
\end{figure}

\refFigure{fig:homotopic_constraints} shows the free space-time envelopes $\bm{\zeta}^{o,\Delta}_{t_p}$ at a single prediction step $t_p=2.4\si{\second}$ for the maneuver envelope $m_{15}$. With the decision to pass the pedestrian on the right, the free-space resembles that by the left boundary going around the predicted occupancy of the pedestrian. The predicted oncoming traffic participant does not intersect with the lane of the ego vehicle and thus with the free-space. Based on the decision to pass the standing vehicle on the left, the right boundary goes of the free-space goes around the obstacle. The merged free space-time maneuver envelope $\bm{\zeta}^{m}_{t_p}$ is displayed below (blue). In cartesian coordinates, it resembles the intersection of all the decision envelopes $\bm{\zeta}^{o_{1-3},\Delta}_{t_p}$. As we operate in Frenet coordinates, we calculate it according to equation \refEquation{equation:maneuver_envelopes}.

\begin{figure}[ht]
\vspace{0.2cm} %
\centering
\subfigure{\label{subfig:a}\includegraphics[width=0.9\columnwidth, frame]{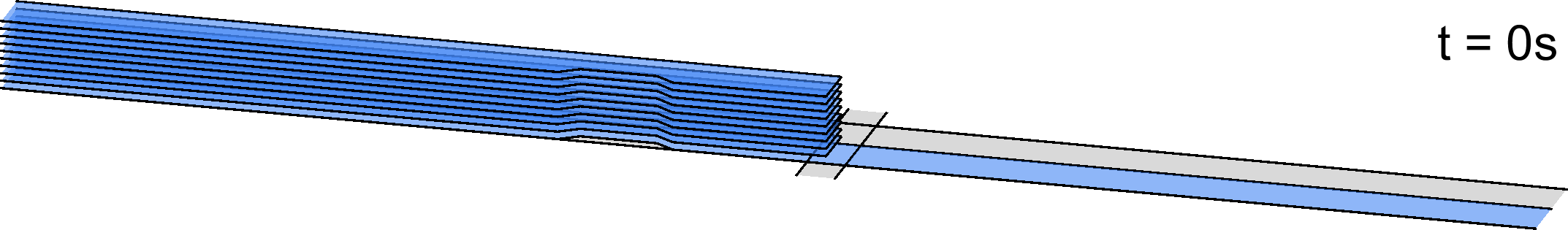}}
\subfigure{\label{subfig:a}\includegraphics[width=0.9\columnwidth, frame]{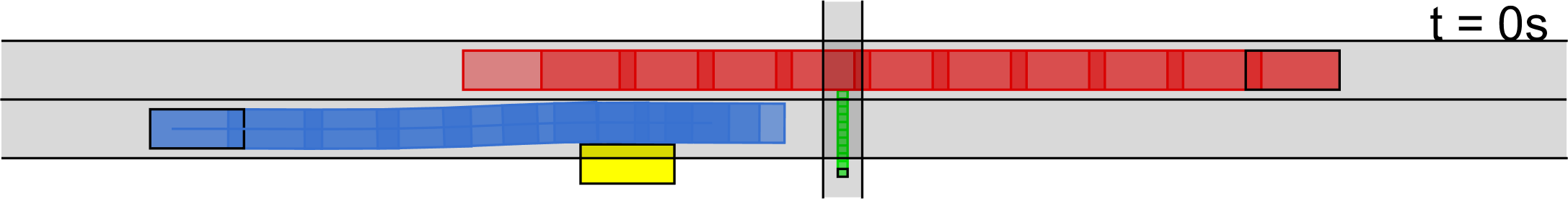}}
\subfigure{\label{subfig:b}\includegraphics[width=0.9\columnwidth, frame]{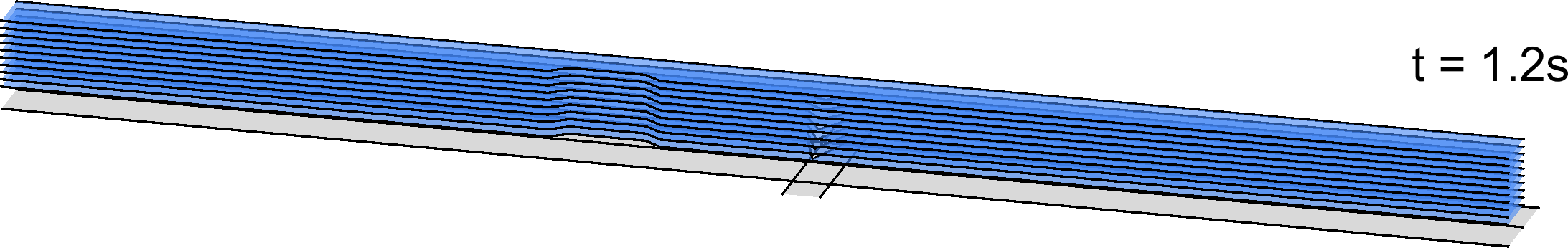}}
\subfigure{\label{subfig:b}\includegraphics[width=0.9\columnwidth, frame]{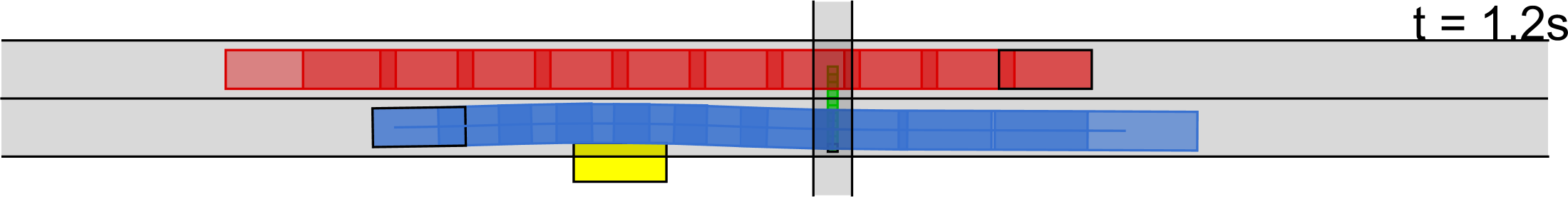}}
\subfigure{\label{subfig:c}\includegraphics[width=0.9\columnwidth, frame]{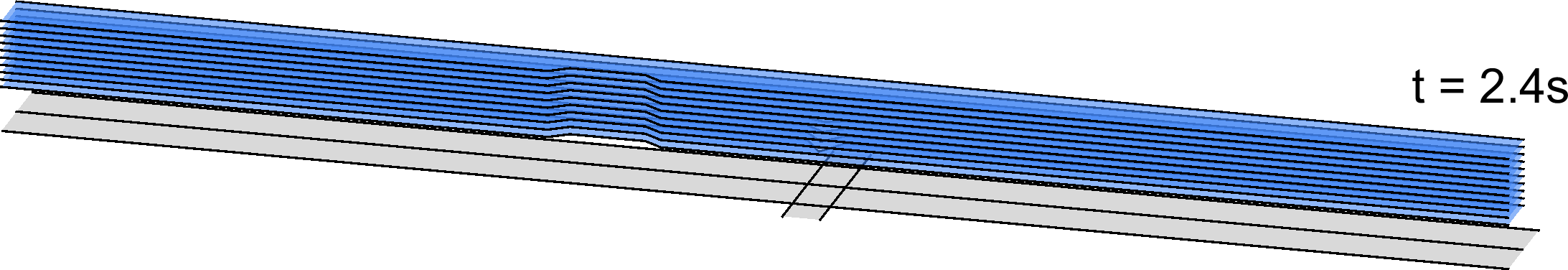}}
\subfigure{\label{subfig:c}\includegraphics[width=0.9\columnwidth, frame]{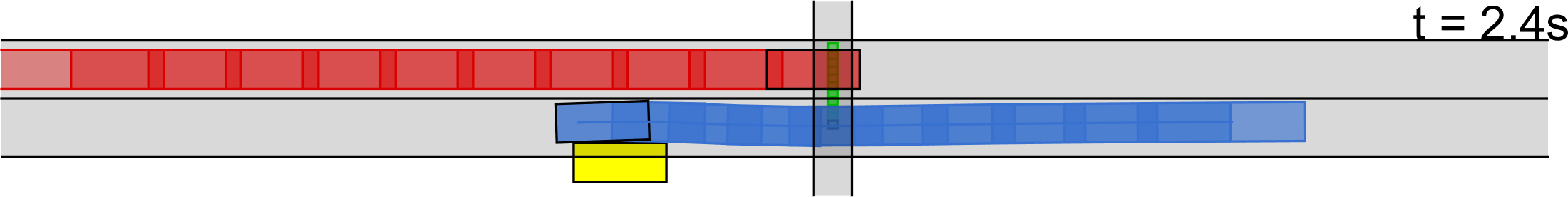}}
\subfigure{\label{subfig:d}\includegraphics[width=0.9\columnwidth, frame]{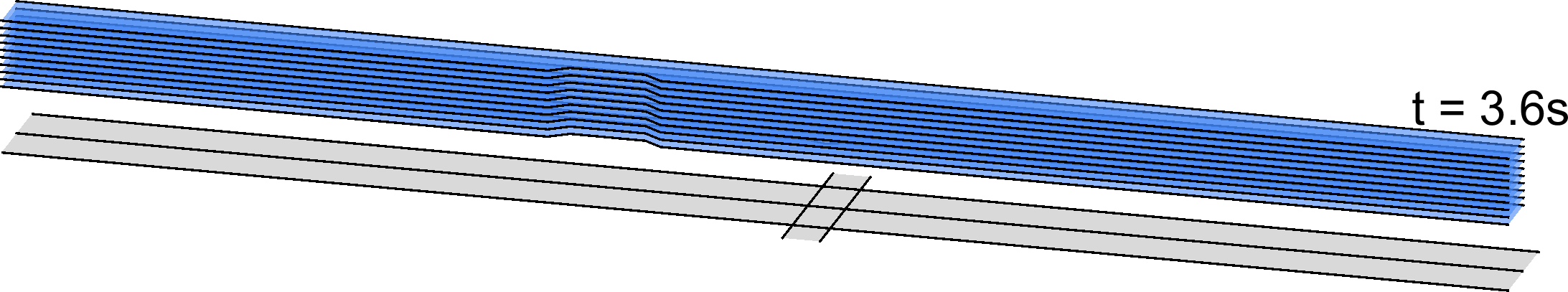}}
\subfigure{\label{subfig:d}\includegraphics[width=0.9\columnwidth, frame]{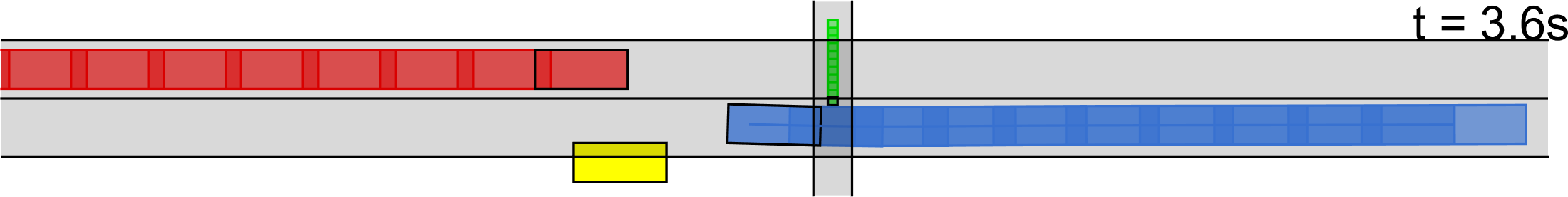}}
\caption{Vehicle movement at $t_0=0\si{\second}$, $t_6=1.2\si{\second}$ , $t_{12}=2.4\si{\second}$ and $t_{18}=3.6\si{\second}$. The optimal free space-time maneuver envelopes $\bm{\zeta}^{m}_{t_p}$ are also displayed for the each prediction time step.}
\label{fig:mpc_pedestrian}
\end{figure}

\refFigure{fig:mpc_pedestrian} shows the outcome of the full simulation over time. At $t_0=0\si{\second}$, the planner chooses to execute $m_{14}$, which means passing after the pedestrian. See \refTable{tab:maneuver_envelope_example} for a full explanation of the maneuver. Note that for $t_0$, $\bm{\zeta}^{m_{14}}_{t_{p1}|t_0}$ includes the full lane length, as the pedestrian will not have entered the lane at $t_{p1}$. For all following prediction steps, the pedestrian has to be taken into account either through longitudinal or lateral decision making. Because of this, the free-space envelopes $\bm{\zeta}^{m_{14}}_{t_{p2...N}|t_0}$ end before the pedestrian.

At $t_6=1.2\si{\second}$, the planner changes from $m_{14}$ to $m_{15}$. $\bm{\zeta}^{m_{15}}_{t_{p1...N}|t_6}$ thus do not stop before the pedestrian, but avoid him on the right, similar to \refFigure{fig:homotopic_constraints}.

The ego vehicle thus avoids the stationary vehicle, goes back to the ego lane and around the pedestrian while maintaining its initial speed. By modifying the optimality criteria, we could select a more defensive driving style.

\begin{table}
\vspace{0.2cm} %
\centering
\caption{Maneuver envelopes $m_{14}$ and $m_{15}$ with decisions $\Delta_j$ for each obstacle $O_i$.}
\begin{tabular}{lccc}
\hline
Maneuver Envelope $m$ & \multicolumn{3}{c}{Decisions $\Delta_j$} \\
\cline{2-4}
& $O_1$ & $O_2$ & $O_3$\\
\hline
$m_{14}$ & after & right & left\\
$m_{15}$ & right & right & left\\
\hline
\end{tabular}
\label{tab:maneuver_envelope_example}
\end{table}

%% file: conclusion.tex
\section{Conclusion and Future Work}
\label{sec:conclusion}
In this work, we proposed a fused trajectory optimization and maneuver selection by introducing a generic decision mechanism to derive maneuver sequences and a semantic language to reason about the maneuver of each obtained trajectory. By separating longitudinal and lateral motion in the trajectory planner, we simplify the constraint formulation as well as the planning problem, which thus allows us to compute multiple trajectories. Note that the maneuver selection framework could be used with other trajectory planners as well. As demonstrated in the simulation results, the novel approach can plan comfortable and safe spatiotemporal trajectories in complex urban driving scenarios. 

The growing number of maneuver types with the number of obstacles still poses a major problem. We will investigate other approaches for the spatiotemporal topological analysis in the future, with the emphasis of discarding infeasible maneuver types. Machine Learning could be used as a heuristic to reduce the number of combinatorial sub-problems. Mixed integer quadratic programming could be investigated to improve the selection of the best trajectory. Other semantic information such as traffic rules could be incorporated into the semantic selection process. Accelerating the computation of the decision envelopes will need to be addressed in future work. We plan to implement the framework in C++ to investigate and improve the real-time capabilities of our approach and to allow for a real-world validation on a full-size research vehicle. In the future, we plan to investigate the possibility of incorporating prediction uncertainty and interaction awareness and the robustness against occlusions.

%% file: appendix.tex
\subsection{Notes on Longitudinal Optimal Control Problem}
The continuous system matrices for the longitudinal planning scheme stated in \Secref{sec_sub:longitudinal_planning} are
\begin{equation}
\bm{A}(t) = \begin{bmatrix}
0 & 1 & 0 & 0 \\
0 & 0 & 1 & 0 \\
0 & 0 & 0 & 1 \\
0 & 0 & 0 & 0 \\
\end{bmatrix}\,,
\end{equation}
\begin{equation}
\bm{B}(t) = \begin{bmatrix}
0 & 0 & 0 & 1\\
\end{bmatrix}
\end{equation}
and
\begin{equation}
\bm{C}(t) = \begin{bmatrix} 
1 & 0 & 0 & 0 \\ 
0 & 1 & 0 & 0 \\ 
0 & 0 & 1 & 0 \\ 
\end{bmatrix}\,.
\end{equation}
\subsection{Notes on Lateral Optimal Control Problem}
For the lateral planning problem described in \Secref{sec_sub:lateral_planning}, the system matrices are
\begin{equation}
\bm{A}(t) = \begin{bmatrix}
0 & v(t) & 0 & -v(t) & 0 \\
0 & 0 & v(t) & 0 & 0 \\
0 & 0 & 0 & 0 & 0 \\
0 & 0 & 0 & 0 & v(t) \\
0 & 0 & 0 & 0 & 0 \\
\end{bmatrix}\,,
\end{equation}

\begin{equation}
\bm{B}(t) = \begin{bmatrix}
0 & 0 & 1 & 0 & 0 \\
\end{bmatrix}\,,
\end{equation}
 
\begin{equation}
\bm{C}(t) = 
\begin{bmatrix} 
1 & 0 & 0 & 0 & 0 \\ 
1 & l/2 & 0 & -l/2 & 0 \\ 
1 & l & 0 & -l & 0 \\ 
0 & 0 & 1 & 0 & v(t) \\ 
\end{bmatrix}
\end{equation}
and
\begin{equation}
\bm{E}(t) = \begin{bmatrix}
0 & 0 & 0 & 0 & 1 \\
\end{bmatrix}\,.
\end{equation}

\subsection{Notes on Batch Formulation for Optimal Control Problem}
The batch matrices are defined as following:
\begin{equation}
\mathcal{A} = 
\begin{bmatrix}
\bm{A}_0^T & \Bigg(\displaystyle\prod_{q=0}^{1} \bm{A}_{1-q}\Bigg)^T & \dots & \Bigg(\displaystyle\prod_{q=0}^{N-1} \bm{A}_{N-1-q}\Bigg)^T \\
\end{bmatrix}
\end{equation}
\begin{equation}
\mathcal{B} = 
\begin{bmatrix}
\bm{B}_0 & 0 & \dots & 0 \\
\bm{A}_1 \bm{B}_0 & \bm{B}_1 & \dots & 0 \\
\vdots & \ddots & \ddots & \vdots \\
\Bigg(\displaystyle\prod_{q=1}^{N-1} \bm{A}_{N+0-q}\Bigg) \bm{B}_0 & \dots & \bm{A}_{N-1} \bm{B}_{N-2} & \bm{B}_{N-1} \\
\end{bmatrix}
\end{equation}

\begin{equation}
\mathcal{C} = 
\begin{bmatrix}
\bm{C}_{1} & & \\
& \ddots & \\
& & \bm{C}_{N}
\end{bmatrix}
\end{equation}

$\mathcal{E}$ is derived similarly to $\mathcal{B}$.